\documentclass[]{memtensor}

\usepackage[toc,page,header]{appendix}
\usepackage[utf8]{inputenc} 
\usepackage[T1]{fontenc}    
\usepackage{hyperref}       
\usepackage{url}            
\usepackage{booktabs}       
\usepackage{amsfonts}       
\usepackage{nicefrac}       
\usepackage{microtype}      
\usepackage[table]{xcolor}  
\usepackage{amsmath} 
\usepackage{amsthm}
\usepackage{etoolbox}
\usepackage{lipsum}  
\usepackage{minitoc}
\usepackage{hyperref}
\usepackage{graphicx}
\usepackage{wrapfig}
\usepackage{arydshln}
\usepackage{multirow} 
\usepackage{booktabs}  
\usepackage{enumerate}
\usepackage{tablefootnote}  
\usepackage{threeparttable}

\usepackage{titletoc}  
\usepackage[table]{xcolor}  
\usepackage{makecell}
\usepackage{colortbl}      
\usepackage[dvipsnames]{xcolor}
\usepackage{multirow, multicol}
\usepackage{arydshln}

\newcommand{\incgreenl}[1]{\textcolor{ForestGreen}{\scriptsize\textbf{#1}}}
\newcommand{\incred}[1]{\textcolor{Red}{\scriptsize\textbf{#1}}}

\newcommand{\incblue}[1]{\textcolor{MidnightBlue}{\textbf{#1}}}
\newcommand{\incgreen}[1]{\textcolor{ForestGreen}{\textbf{#1}}}
\definecolor{lightgray}{RGB}{242,242,242}
\definecolor{lightblue}{RGB}{230,242,248}

\title{
\raisebox{-2.0em}{
  \parbox[t]{0.5in}{\includegraphics[width=0.8in]{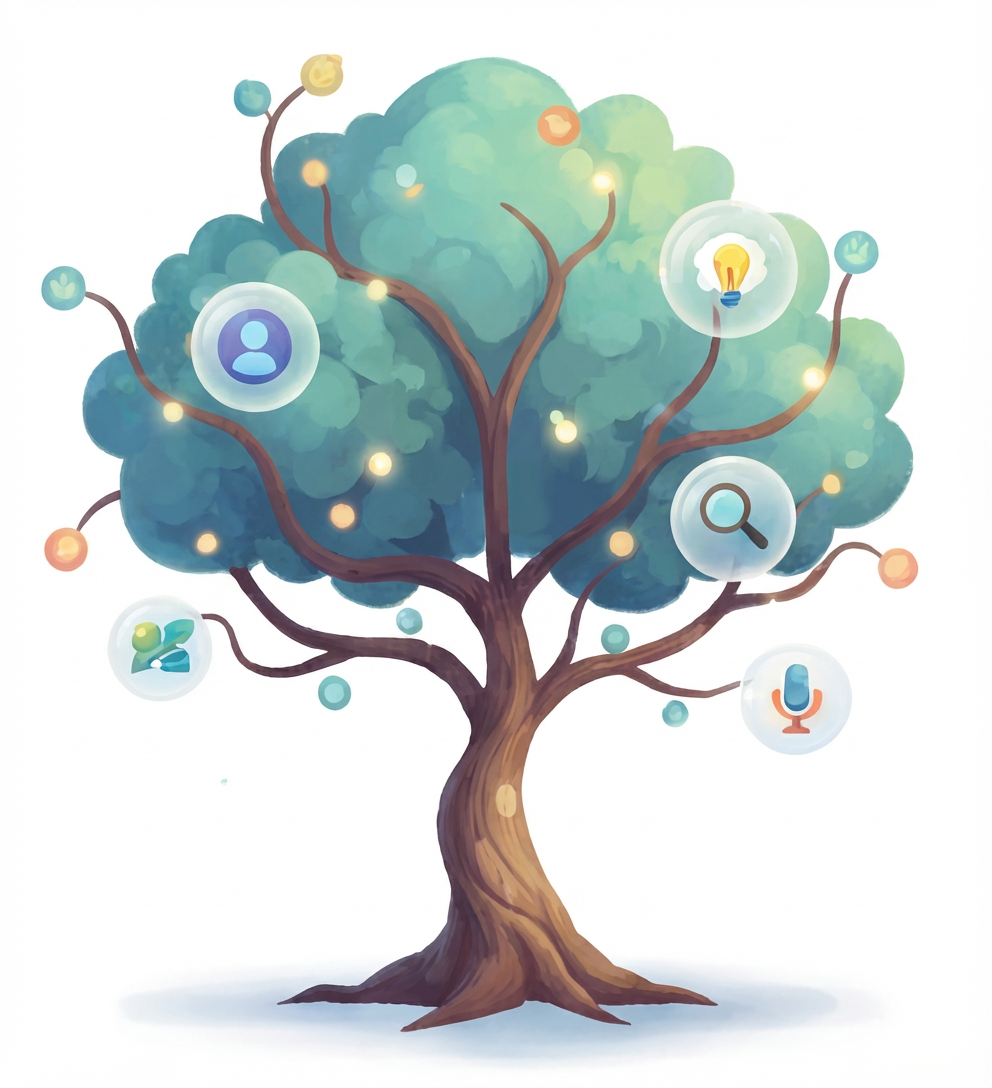}} 
  }
\begin{tabular}[t]{l} 
  \parbox[t]{0.8\textwidth}{\centering \LARGE
    Inside Out: \\Evolving User-Centric Core Memory Trees\\ for Long-Term Personalized Dialogue Systems
  }
\end{tabular}}

\author[1,2,3]{Jihao Zhao} 
\author[4]{Ding Chen}
\author[2,5]{Zhaoxin Fan}
\author[2,3]{Kerun Xu}
\author[2,6]{Mengting Hu}

\author[2,3]{Bo Tang}
\author[2,3]{Feiyu Xiong}
\author[2,3,\dag]{Zhiyu Li}

\affiliation[1]{School of Information, Renmin University of China}
\affiliation[2]{MemTensor (Shanghai) Technology Co., Ltd.}
\affiliation[3]{Institute for Advanced Algorithms Research, Shanghai}
\affiliation[4]{China Telecom Research Institute}
\affiliation[5]{Beijing University of Aeronautics and Astronautics}
\affiliation[6]{Nankai University}

\abstract{
Existing long-term personalized dialogue systems struggle to reconcile unbounded interaction streams with finite context constraints, often succumbing to memory noise accumulation, reasoning degradation, and persona inconsistency. To address these challenges, this paper proposes \textbf{Inside Out}, a framework that utilizes a globally maintained \textbf{PersonaTree} as the carrier of long-term user profiling. By constraining the trunk with an initial schema and updating the branches and leaves, PersonaTree enables controllable growth, achieving memory compression while preserving consistency. Moreover, we train a lightweight \textbf{MemListener} via reinforcement learning with process-based rewards to produce structured, executable, and interpretable \(\{\texttt{ADD}, \texttt{UPDATE}, \texttt{DELETE}, \texttt{NO\_OP}\}\) operations, thereby supporting the dynamic evolution of the personalized tree. During response generation, PersonaTree is directly leveraged to enhance outputs in latency-sensitive scenarios; when users require more details, the agentic mode is triggered to introduce details on-demand under the constraints of the PersonaTree. Experiments show that PersonaTree outperforms full-text concatenation and various personalized memory systems in suppressing contextual noise and maintaining persona consistency. Notably, the small MemListener model achieves memory-operation decision performance comparable to, or even surpassing, powerful reasoning models such as DeepSeek-R1-0528 and Gemini-3-Pro. 
}

\correspondence{Team Leader at \email{lizy@memtensor.cn}}
\checkdata[Author Legend]{\dag{}Corresponding author}
\checkdata[Code]{\url{https://github.com/MemTensor/InsideOut}}

\begin{document}
\maketitle

\section{Introduction}
\begin{quote}
  \textit{Core memories shape Riley's personality islands, with each island serving as a unique emblem of her identity.}\\[-5pt]
   \verb|                                                                  |------  \textit{"Inside Out"}
   \vspace{-5pt}
\end{quote}

With the rapid advancement of large language models (LLM), dialogue-based agents have demonstrated substantial potential in applications such as personal assistants, affective companionship, and long-term question answering~\citep{chhikara2025mem0,rasmussen2025zep,li2025memos_long}. However, within personalized dialogue systems aimed at fostering long-term human-machine trust and emotional connection, a fundamental contradiction exists between the finite context window and the unbounded growth of interaction history~\citep{xiao2024efficient,liu2024lost}. As conversational turns continue to accumulate, the traditional single-context paradigm encounters a severe form of context saturation: indiscriminate aggregation of massive historical information not only drives computational costs sharply upward, but also introduces substantial irrelevant noise, markedly degrading the signal-to-noise ratio. More critically, this unstructured accumulation makes it difficult for the model to accurately extract and sustain a user’s personal characteristics from lengthy histories, leading to personalization inconsistency over long-term interactions and thereby seriously undermining user experience and the system’s long-term usability~\citep{zhong2024memorybank,salemi2024lamp}.

To address these challenges, existing studies have primarily explored routes such as explicit profile augmentation and vector-based retrieval, yet neither directly confronts the central bottleneck of personalized memory evolution. Profile-based approaches rely on predefined, static attributes; they are not only slow to update but also struggle to capture implicit cues that users reveal over prolonged interactions, including linguistic style, deeper value orientations, and affective preferences, resulting in superficial personalization modeling~\citep{tan2023usermodeling}. In contrast, memory-augmented agents based on vector retrieval, while introducing external storage, still essentially treat memory as text fragments or simple lists of facts. Such systems lack an intrinsic, trained decision mechanism for determining which information merits long-term retention, and instead often depend on rigid heuristics or elaborate prompt engineering~\citep{liu2023thinkinmemory}. This accumulation of memories without value-based judgment causes the memory repository either to become bloated and uninterpretable due to noise accretion, or to lose the long-range logical thread through fragmentation of key context, ultimately failing to sustain a vivid and coherent persona~\citep{yoran2024making}.

This discrepancy between memory accumulation and core persona formation” motivates us to return to the foundations of human cognition for an answer. As illustrated by the film "\textit{Inside Out}", individual identity does not stem from a simplistic stacking of all experiences, but rather is constructed upon core memories that shape distinct "Islands of Personality". This aligns with theoretical findings in cognitive psychology, such as Self-Schema theory~\citep{Markus1977SelfschemataAP,tikka2019tailoring}, which emphasizes that humans maintain a stable self-concept by filtering and hierarchically organizing key memories. 

Inspired by these insights, we propose the \textbf{Inside Out} framework, which aims to grow an evolvable user core memory tree "from the inside out" through unbounded interactions. Firstly, to delineate the theoretical boundaries of the memory tree, we construct a hierarchical Schema based on the Biopsychosocial model, scientifically decomposing user characteristics into three core dimensions. This interdisciplinary Schema design establishes the initial structure of the user \textbf{PersonaTree}. Secondly, to endow the system with dynamic evolution, we propose an iterative tree-update mechanism and introduce a reinforcement learning (RL) strategy based on process rewards to train a lightweight model, \textbf{MemListener}. This model learns to compress a continuous stream of unstructured dialogue in real time into standardized tree-structured operations, encoding user core features within the branch and leaf nodes. Finally, addressing the trade-off between efficiency and effectiveness during the inference stage, this paper designs an adaptive response generation mechanism: In latency-sensitive scenarios, a fast mode is enabled to perform reasoning directly based on the PersonaTree. When facing long-tail detail requirements, the system switches to the agentic recall mode, utilizing the PersonaTree to guide deep retrieval. The primary contributions of our work are summarized as:

\begin{itemize}
    \item We propose PersonaTree, grounded in the biopsychosocial schema. By transforming unstructured dialogue streams into standardized atomic tree operations in real-time, PersonaTree achieves the dynamic compression, explicit management, and high signal-to-noise ratio maintenance of implicit user profiles.
    \item We design a training strategy utilizing RL with process rewards. Leveraging the constructed dataset of 28k instructions, we train a lightweight model, MemListener, to execute precise memory editing.
    \item Our experiments reveal the potential of a collaborative paradigm where "small models maintain memory while LLMs handle generation". Results show that MemListener achieves memory-decision performance comparable to strong reasoning models, and that PersonaTree offers a new pathway toward low-cost, highly reliable deployment of long-term personalized dialogue systems.
\end{itemize}

\begin{figure*}[t]
    \centering
    \includegraphics[width=\textwidth]{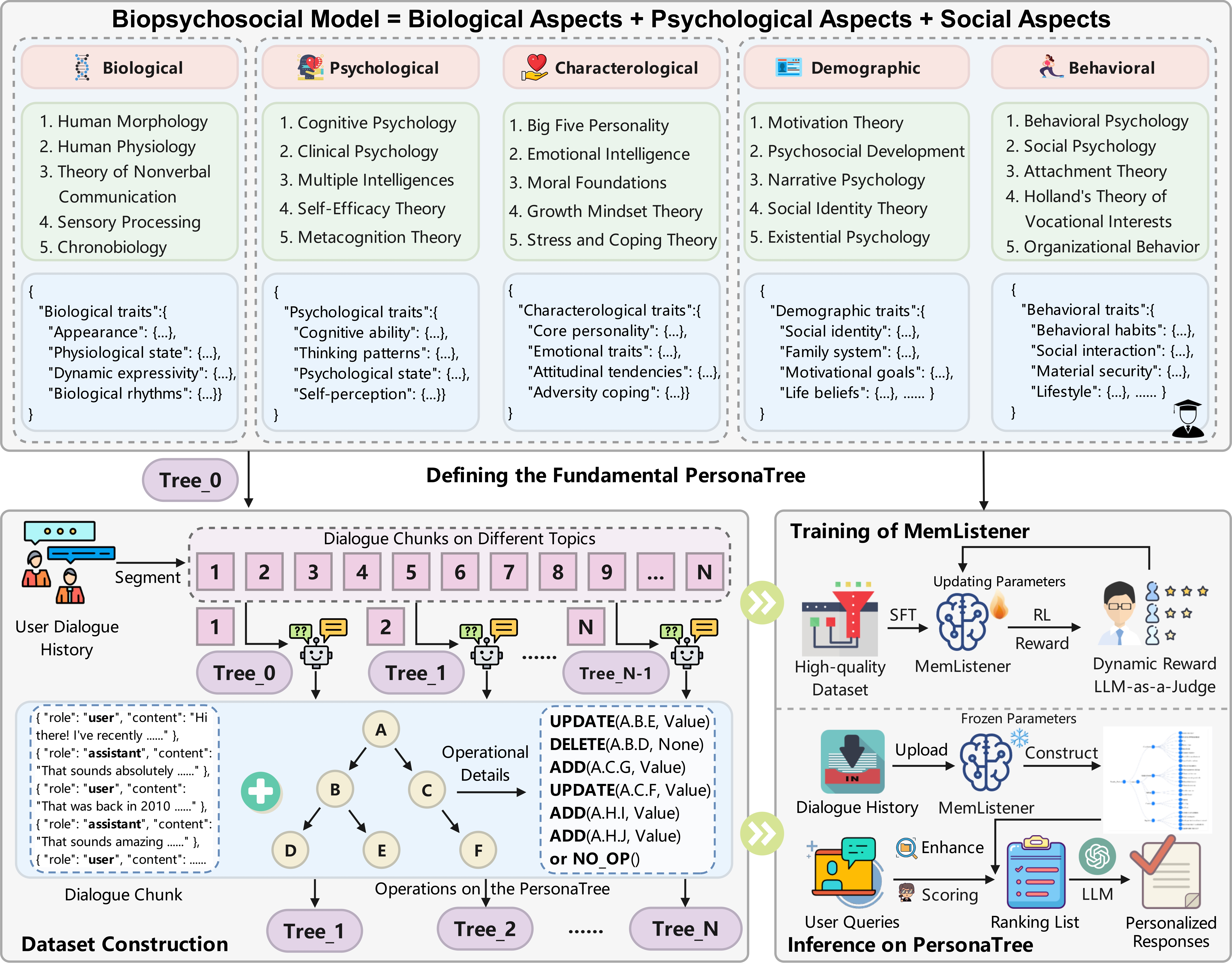}
    \caption{Overview of the entire process of our Inside Out framework.}
    \label{fig:kuangjia}
\end{figure*}

\section{Related Works}
\subsection{Personalization and Memory}
Personalization aims to adapt a dialogue system's linguistic style and interaction policy to a specific user's stable traits and evolving state. In interactive settings, personalization is inherently coupled with memory: models must distill reusable user representations from past interactions and fuse them during generation. \citet{li2016persona} proposed persona-based dialogue generation to mitigate inconsistency and lack of personality in open-domain dialogue, and \citet{zhang2018personalizing} formalized the PersonaChat task. Subsequent studies emphasized multi-dimensional user attributes. For example, \citet{zheng2019personalized} introduced the large-scale multi-turn dataset PersonalDialog. In parallel, \citet{madotto2019personalizing} framed personalization as a meta-learning problem to enable few-shot adaptation. In the LLM era, \citet{chen2024recent} systematically reviewed major directions in personalized dialogue generation, while \citet{tan-etal-2024-democratizing} assigned parameter-efficient personalization modules to users to improve multi-task personalization.

\subsection{LLM Agents with External Memory}
To overcome the limitations of LLMs' finite context windows and endow them with capabilities for continuous learning and long-term interaction, constructing memory systems has emerged as a pivotal research direction. LangMem \footnote{\url{https://github.com/langchain-ai/langmem}} enables continual learning and cross-session consistency by decoupling hot-path memory primitives from backend asynchronous integration. Mem0 \citep{chhikara2025mem0} adopts a multi-level memory architecture to support multi-session retrieval and personalization at relatively low overhead. A-Mem \citep{xu2025mem} builds an evolvable memory network via self-organizing indexing and linking mechanisms. MemoryOS \citep{kang2025memory} manages short, medium, and long-term memory through OS-style hierarchical storage together with corresponding update and retrieval policies to preserve contextual coherence.

\section{The Inside Out Framework}
\subsection{Overview Architecture}
This study proposes the Inside Out Framework, which aims to address the challenges of personalized consistency and contextual forgetting in long-term dialogues through a structured memory evolution mechanism.

\paragraph{Framework Pipeline.}
As shown in Figure~\ref{fig:kuangjia}, the framework consists of three key modules: Dynamic PersonaTree Evolution (Section \ref{Dynamic PersonaTree Evolution}), MemListener Training (Section \ref{section MemListener Training}), and Adaptive Response Generation (Section \ref{Adaptive Response Generation}). First, \textbf{PersonaTree and dataset construction} initializes a persona tree based on the Biopsychosocial Model, segments the user's dialogue history into consecutive dialogue chunks, and generates operations on the PersonaTree, thereby constructing a memory evolution dataset. Second, \textbf{MemListener training} leverages the resulting high-quality dataset to update the parameters of MemListener via supervised fine-tuning (SFT) and RL with a dynamic reward mechanism, enabling it to extract structured memory from unstructured dialogues. Finally, \textbf{PersonaTree inference} freezes the MemListener parameters at application time, reconstructs the attribute tree from the dialogue history, uses this structured memory to enhance user queries, and ultimately generates personalized responses through an LLM.

\paragraph{Problem Formulation.}
We define the task of a personalized dialogue system as a process of maximizing response utility over an infinitely long dialogue stream. Given a user $U$ with a historical dialogue sequence $H = \{x_1, y_1, ..., x_t, y_t\}$, where $x$ denotes user inputs and $y$ denotes system responses, conventional context-window approaches attempt to directly model $P(y_t \mid H_{t-k:t})$, but are constrained by the window length $k$. Our framework introduces an explicit, structured user state $\mathcal{T}$ (i.e., PersonaTree), thereby reformulating the problem as state tracking and state-conditioned generation. The goal is to learn a state update function $f_{\text{update}}$ such that:
\begin{equation}
\mathcal{T}_t = f_{\text{update}}(\mathcal{T}_{t-1}, D_t)
\end{equation}
\begin{equation}
y_t = f_{\text{gen}}(x_t, \mathcal{T}_t, f_{\text{recall}}(\mathcal{T}_t,H))
\end{equation}
where $D_t$ denotes the current dialogue chunk, $f_{\text{gen}}$ produces the system reply given the current user input and the tracked user state, and $f_{\text{recall}}$ is a retrieval function that recalls relevant historical snippets from the full dialogue history $H$ conditioned on the current state $\mathcal{T}_t$.

\subsection{Dynamic PersonaTree Evolution}
\label{Dynamic PersonaTree Evolution}
\paragraph{PersonaTree Initialization.}
At system startup, we construct an initial PersonaTree to serve as the starting point of the user's long-term structured state. Specifically, we first determine the set of writable trunk and leaf fields according to a predefined unified schema, and constrain the storage type of each leaf node to a descriptive string, which is used to hold a compressed summary of the user's core personalized attributes. This design ensures that memory capacity remains controllable and prevents unbounded growth as the dialogue progresses. The schema is informed by interdisciplinary human-factors and psychological theory frameworks, with its theoretical grounding illustrated in Figure~\ref{fig:kuangjia}. Subsequently, under the schema constraints, we initialize the leaf nodes (allowing empty strings or default placeholder text), thereby obtaining the initial persona tree $\mathcal{T}_0$. The specific initial PersonaTree instance adopted in this paper is provided in Appendix~\ref{Initial PersonaTree Instance}.

\paragraph{Iterative PersonaTree Updating.}
To enable scalable maintenance of long-term personalized memory over an infinitely long dialogue stream, we adopt an iterative updating mechanism: any input modality (historical file import, short-snippet input, or real-time cache triggering) is normalized into a dialogue-chunk sequence $(D_1,\dots,D_N)$, and for each $D_t$ we execute a closed-loop update of operation-list generation, safe parsing and execution, versioned persistence.

\textbf{Step 0: System Loading.}
The system loads the text fields of all leaf nodes, yielding the initial state $\mathcal{T}_0$. Meanwhile, the task specification and system constraints are abstracted into a rule set $\mathcal{R}$, including update rules, writable scope, leaf constraints, and output format.

\textbf{Step $t=1,\dots,N$.}
For any dialogue chunk $D_t$, the system executes the following three stages:

\textbf{(a) State Construction.}
Given a dialogue chunk $D_t$, set $\mathrm{Input}_t \leftarrow D_t,\mathcal{T}_{t-1}$. 

\textbf{(b) Operation List Generation.}
Conditioned on $(D_t,\mathcal{T}_{t-1},\mathcal{R})$, LLM outputs an operation list $\mathcal{O}_t$, consisting of one or more atomic operations that strictly follow a predefined operation grammar. The operation types are limited to:
\begin{itemize}
  \item \texttt{ADD(path, value)}: write descriptive text to the specified path; if the path does not exist, it may be created under an extended-schema policy;
  \item \texttt{UPDATE(path, value)}: perform an overwrite rewrite on the target leaf node, updating its text to the new value;
  \item \texttt{DELETE(path, value)}: clear the target leaf node or write a deletion marker to indicate that this type of information should be removed from long-term memory;
  \item \texttt{NO\_OP()}: the current dialogue chunk does not contain stable core persona information that should be written to the PersonaTree.
\end{itemize}
For update operations, our framework unifies them as rewrites of leaf strings. More importantly, potential conflicts between new and old information are resolved by LLM during the generation of $\mathcal{O}_t$. Based on $D_t$ and the contextual old values in $\mathcal{T}_{t-1}$, the model must decide whether to overwrite prior information, preserve salient change cues, or solely append new information. In other words, conflict resolution is explicitly lifted to the policy-generation stage, so as to leverage the LLMs' holistic inference over semantics, temporal order, and narrative consistency.

\textbf{(c) Parsing and Execution.}
This module serves as a safety gate that enforces structural and capacity constraints when applying $\mathcal{O}_t$: it validates that each path targets a permissible leaf, ensures each value is a string or an allowed deletion marker to avoid parsing ambiguity and state pollution, and applies length control by compressing any overlong value to satisfy the per-leaf budget. Importantly, it performs no conflict resolution or secondary semantic rewriting; it only executes the prescribed operations under these constraints.

\textbf{(d) Versioned Persistence.}
After the execution, the updated tree state $\mathcal{T}_t$ is materialized and persisted as a new version, either serialized to a JSON file or stored in a JSON-capable database (e.g., document stores such as MongoDB). Iterating over $t=1,\dots,N$ yields a traceable evolution sequence $\{\mathcal{T}_0,\dots,\mathcal{T}_N\}$, with $\mathcal{T}_N$ serving as the compressed long-term memory for retrieval-augmented and personalized generation at the final task-query stage.

\subsection{MemListener Training}
\label{section MemListener Training}
\paragraph{Training Data Synthesis.}
During training data construction, we select subsets from HaluMem \citep{chen2025halumem} and PersonaMem \citep{jiang2025know} that are relevant to implicitly characterizing user-specific attributes as the raw corpus sources. Using the dynamic PersonaTree evolution procedure described in Section~\ref{Dynamic PersonaTree Evolution} as the backbone, we invoke DeepSeek-R1-0528 to generate supervision signals for training. 

\paragraph{Warm-up via SFT.}
We first perform full supervised fine-tuning to initialize the base model as a MemListener that can stably generate operation lists. For any training sample, let the input context be $s$ (including the dialogue chunk, the previous tree state, and rule constraints), and the target output be $o$ (the ground-truth operation sequence segment). We optimize a standard autoregressive cross-entropy objective:
\[
\mathcal{L}_{\mathrm{SFT}}(\theta) = -\frac{1}{\tau}\sum_{t=1}^{\tau}\log P_\theta\!\left(o_t \mid o_{<t}, s\right),
\]
where $\theta$ denotes the model parameters and $\tau$ is the target sequence length. The goal of this stage is to train the model to make executable and traceable update decisions within strict grammar and write boundaries, ensuring that the generated operation sequences are structurally stable and usable, thus providing a reliable starting point for subsequent alignment.

\paragraph{Alignment via Process-Reward RL.}
After the SFT warm-up, we continue alignment with the remaining data using RL driven by process-based rewards. We set the model's maximum context length to 11K tokens, with the input length capped at 10K, which constitutes a typical ultra-long sequence optimization setting. Therefore, we enable the DAPO loss for process-reward alignment within a GRPO framework. We perform credit assignment via token-level policy gradients, and mitigate entropy collapse by decoupling the clipping range from dynamic sampling constraints, ensuring that the within-group advantage retains non-zero variance. This makes the method better suited to long-chain reasoning and structured operation-sequence generation. In addition, by limiting the maximum generation length and applying dynamic resampling to filter degenerate groups, we reduce training noise introduced by truncation and within-group advantage degeneration. 

Concretely, for each training sample, we take $s$ (including the dialogue chunk, the previous tree state, and update constraints) as input, and use the manually verified reference output $y^\star$ as the ground-truth operation trajectory. During optimization, we sample a group of candidate outputs $\{y_i\}_{i=1}^{G}$ from the old policy $\pi_{\theta_{\text{old}}}(\cdot\mid s)$, compute sequence-level returns under the process reward function $R(\cdot)$ as $R_i = R(y_i, y^\star; s)$, and then update the policy parameters $\theta$ using the DAPO objective:
\begin{equation*}
\begin{aligned}
\mathcal{J}_{\mathrm{RL}}(\theta)
=\ & \mathbb{E}_{(s,y^\star)\sim \mathcal{D},\ \{y_i\}_{i=1}^{G}\sim \pi_{\theta_{\mathrm{old}}}(\cdot\mid s)}
\Bigg[
\frac{1}{\sum_{i=1}^{G} |y_i|}
\sum_{i=1}^{G}\sum_{t=1}^{|y_i|}
\min\Big(
r_{i,t}(\theta)\,\hat{A}_{i,t},\
\operatorname{clip}\!\big(r_{i,t}(\theta),     \,1-\varepsilon_{\mathrm{low}},\,1+\varepsilon_{\mathrm{high}}\big)\,\hat{A}_{i,t}
\Big)
\Bigg],
\label{eq:memlistener_dapo_objective}
\end{aligned}
\end{equation*}

\begin{equation*}
\text{s.t.}~~
0<
\Bigl|\{\,y_i \mid \operatorname{is\_equivalent}(y^\star, y_i)\,\}\Bigr|
< G.
\end{equation*}
Policy updates are based on the importance ratio
\begin{equation*}
\begin{aligned}
r_{i,t}(\theta)
=
\frac{
\pi_{\theta}\!\left(y_{i,t}\mid s, y_{i,<t}\right)
}{
\pi_{\theta_{\mathrm{old}}}\!\left(y_{i,t}\mid s, y_{i,<t}\right)
},~~
\hat{A}_{i,t}
=
\frac{
R_i-\operatorname{mean}\!\left(\{R_j\}_{j=1}^{G}\right)
}{
\operatorname{std}\!\left(\{R_j\}_{j=1}^{G}\right)
},
\label{eq:memlistener_dapo_advantage}
\end{aligned}
\end{equation*}
where $R_i\in[-1,1]$ denotes the sequence-level score assigned by the evaluator for the $i$-th sampled output $y_i$ in a group of size $G=8$. We apply within-group standardization to obtain the advantage estimate $\hat{A}_{i,t}$, thereby improving training stability. We adopt $\varepsilon_{\mathrm{low}}=0.2$ and $\varepsilon_{\mathrm{high}}=0.28$ to relax the upper-bound clipping, providing greater update headroom for increasing the probabilities of low-probability exploratory tokens. The detailed data construction, training parameter settings, and the design of the reward function are presented in Appendix \ref{MemListener Training}.

\subsection{Adaptive Response Generation}
\label{Adaptive Response Generation}
During the inference, we treat the final tree state $\mathcal{T}_N$ as structured long-term memory and adopt an adaptive response strategy to satisfy both low-latency and high-coverage requirements.
\paragraph{PersonaTree-Augmented Generation.}
For the latency-sensitive interaction scenario, the system enables a lightweight fast mode: it directly reads out the structure of $\mathcal{T}_N$ along with the non-empty leaf texts as a personalized prior, concatenates them with the user query $q$ as the input context, and generates an answer in a single pass. 

\paragraph{Agentic Recall and Fusion.}
When the user explicitly requests additional details or the query exhibits stronger long-tail characteristics, the system switches to the agentic recall mode. Concretely, we generate a set of expanded queries $\{\tilde{q}^{(k)}\}_{k=1}^{K}$ from the original query $q$ conditioned on $\mathcal{T}_N$, where each $\tilde{q}^{(k)}$ emphasizes a different attribute dimension or potential missing aspect relevant to the question. The system retrieves candidate evidence sets $\{d^{(k)}_j\}$ in parallel for $\{q^{(k)}\}$, and reranks them based on relevance to obtain a fused context $C$. Finally, we generate the final answer conditioned on $[q,\,\mathcal{T}_N,\,C]$. This procedure operates under a gated policy that is triggered only when necessary, improving answer quality in complex scenarios while keeping the overall interaction cost controllable.

\section{Experiments}
\subsection{Experimental Setup}
\paragraph{Datasets and Metrics.}
We conduct experiments on the PersonaMem benchmark \citep{jiang2025know}. This dataset is centered on user personas: each instance contains a user's static demographic information as well as dynamic attributes that evolve over time. For interaction-history construction, each history consists of approximately 10 multi-turn conversations concatenated in chronological order, resulting in an overall context length of about 32k tokens and covering 15 categories of real-world tasks that require personalization. For evaluation, we use accuracy as the primary metric, reporting overall accuracy and further breaking it down into per-skill accuracies across seven query-skill categories. Under the discriminative setting, the model directly outputs the selected option.

\paragraph{Baselines.}
To evaluate the effectiveness of the proposed method, we compare against two standard interaction paradigms Only LLM and Full Context as well as four representative memory-management frameworks LangMem, Mem0 \citep{chhikara2025mem0}, A-Mem \citep{xu2025mem}, and MemoryOS \citep{kang2025memory}. For each baseline, we use the best-performing configurations reported in the original works (e.g., agentic retrieval). Detailed descriptions and experimental setups are provided in Appendix~\ref{add_Baselines}.

\paragraph{Implementation Details.} 
In our approach, the construction of core training data leverages the DeepSeek-R1-0528. To encourage diverse generations, we set temperature to 0.7 and top-p to 0.9. For retrieval components, both our method and all baseline systems share the same settings: we use BGE-M3 as the retriever and BGE-Reranker-Large as the re-ranking model. We set the retrieval count to 4 and ensure that different methods maintain a fundamentally consistent length of retrieval context. All training and evaluation experiments are conducted on a single node equipped with 8 NVIDIA H200-141GB GPUs.

\subsection{Main Results}
\begin{table*}[ht!]
    \centering
    \resizebox{\linewidth}{!}{
    \begin{threeparttable}
        \begin{tabular}{lcccccccc}
        \toprule
        \textbf{Methods} & \textbf{Overall} & \textbf{Recall-Facts} & \textbf{Pref-Rec} & \textbf{New-Ideas} & \textbf{Recall-Reason} & \textbf{Pref-Evol} & \textbf{Gen-New} & \textbf{Recall-User} \\ 
        \midrule
        \multicolumn{9}{l}{\texttt{\incblue{DeepSeek-V3.1 (Response)}}} \\
        \quad Only LLM & 52.63 & 59.69 & 43.64 & 5.38 & 80.81 & 65.47 & 42.11 & 52.94 \\
        \quad ALL Dialogue & 63.84 & 78.29 & 61.82 & 12.90 & 81.82 & 65.47 & 77.19 & 76.47 \\
        \quad LangMem\incgreenl{$+$DeepSeek-R1-0528} & 57.05 & 70.54 & 43.64 & 11.83 & 80.81 & 58.99 & 63.16 & 70.59 \\
        \quad Mem0\incgreenl{$+$DeepSeek-R1-0528} & 60.44 & 79.84 & 52.73 & 5.38 & 83.84 & 61.15 & 64.91 & 82.35 \\
        \quad A-Mem\incgreenl{$+$DeepSeek-R1-0528} & 59.76 & 79.84 & 54.55 & 4.30 & 86.87 & 56.83 & 64.91 & 76.47 \\
        \quad MemoryOS\incgreenl{$+$DeepSeek-R1-0528} & 62.48 & 72.87 & \textbf{65.45} & 11.83 & 84.85 & 63.31 & 73.68 & 76.47 \\
        \rowcolor{lightblue}
        \quad PersonaTree-ALL\incgreenl{$+$DeepSeek-R1-0528} & \underline{71.14} & \underline{88.37} & 56.36 & \underline{22.58} & \underline{88.89} & 71.22 & \textbf{89.47} & \underline{88.24} \\
        \rowcolor{lightblue}
        \quad PersonaTree-ALL\incgreenl{$+$Qwen2.5-7B-RL} & \textbf{71.31} & \textbf{89.15} & 61.82 & \textbf{23.66} & \textbf{91.92} & 69.06 & \underline{85.96} & 76.47  \\
        \rowcolor{lightblue}
        \quad PersonaTree-ALL\incgreenl{$+$Qwen3-8B-RL} & 70.80 & 84.50 & 61.82 & 21.51 & 87.88 & \underline{72.66} & \underline{85.96} & \textbf{100.00}  \\
        \addlinespace[1pt]
        \rowcolor[rgb]{0.94,0.94,0.94} \quad \qquad w/  PersonaTree$+$Router & 70.12 & 82.17 & \underline{63.64} & 19.35 & \underline{88.89} & \textbf{73.38} & 84.21 & 94.12 \\
        \rowcolor[rgb]{0.94,0.94,0.94} \quad \qquad w/ PersonaTree & 61.97 & 75.19 & 60.00 & 4.30 & 83.84 & 64.75 & 75.44 & 88.24 \\
        \addlinespace[1pt]
        \cdashline{1-9}
        \addlinespace[1pt]
        \multicolumn{9}{l}{\texttt{\incblue{Longcat-Flash-Chat (Response)}}} \\
        \quad Only LLM & 54.33 & 62.02 & 43.64 & 8.60 & 84.85 & \underline{71.22} & 28.07 & 52.94 \\
        \quad ALL Dialogue & 61.80 & 79.07 & 63.64 & 15.05 & 87.88 & 53.96 & 68.42 & 70.59 \\
        \quad LangMem\incgreenl{$+$DeepSeek-R1-0528} & 58.23 & 75.97 & 47.27 & 8.60 & 82.83 & 64.03 & 50.88 & 64.71 \\
        \quad Mem0\incgreenl{$+$DeepSeek-R1-0528} & 59.59 & 80.62 & 52.73 & 5.38 & 85.86 & 62.59 & 49.12 & 76.47 \\
        \quad A-Mem\incgreenl{$+$DeepSeek-R1-0528} & 60.95 & 82.17 & 54.55 & 8.60 & 85.86 & 62.59 & 56.14 & 64.71 \\
        \quad MemoryOS\incgreenl{$+$DeepSeek-R1-0528} & 65.03 & 77.52 & \textbf{72.73} & 11.83 & 86.87 & 66.91 & 70.18 & 76.47 \\
        \rowcolor{lightblue}
        \quad PersonaTree-ALL\incgreenl{$+$DeepSeek-R1-0528} & 72.67 & 88.37 & \textbf{72.73} & 26.88 & \underline{92.93} & 69.78 & \underline{85.96} & 64.71 \\
        \rowcolor{lightblue}
        \quad PersonaTree-ALL\incgreenl{$+$Qwen2.5-7B-RL} & \underline{73.34} &  \textbf{93.02} & 67.27 & \underline{27.96} & \underline{92.93} & 67.63 & \textbf{89.47} & 70.59  \\
        \rowcolor{lightblue}
        \quad PersonaTree-ALL\incgreenl{$+$Qwen3-8B-RL} & \textbf{75.38} & \textbf{93.02} & \underline{70.91} & \textbf{30.11} & \textbf{93.94} & \textbf{71.94} & \underline{85.96} & \textbf{88.24}  \\
        \addlinespace[1pt]
        \rowcolor[rgb]{0.94,0.94,0.94} \quad \qquad w/  PersonaTree$+$Router & 71.82 & \underline{89.92} & \underline{70.91} & 22.58 & 90.91 & 68.35 & 84.21 & \underline{82.35} \\
        \rowcolor[rgb]{0.94,0.94,0.94} \quad \qquad w/ PersonaTree & 65.20 & 79.84 & 67.27 & 10.75 & 87.88 & 68.35 & 68.42 & 76.47 \\
        \addlinespace[1pt]
        \cdashline{1-9}
        \addlinespace[1pt]
        \multicolumn{9}{l}{\texttt{\incblue{DeepSeek-R1-0528 (Response)}}} \\
        \quad Only LLM & 44.14 & 30.23 & 38.18 & 19.35 & 76.77 & 55.40 & 38.60 & 41.18 \\
        \quad ALL Dialogue & 64.86 & 69.77 & 63.64 & 11.83 & 84.85 & 73.38 & 84.21 & 70.59 \\
        \quad LangMem\incgreenl{$+$DeepSeek-R1-0528} & 54.84 & 63.57 & 61.82 & 15.05 & 78.79 & 56.83 & 50.88 & 41.18 \\
        \quad Mem0\incgreenl{$+$DeepSeek-R1-0528} & 49.41 & 32.56 & 56.36 & 24.73 & 81.82 & 56.83 & 43.86 & 58.82 \\
        \quad A-Mem\incgreenl{$+$DeepSeek-R1-0528} & 47.37 & 30.23 & 45.45 & 22.58 & 77.78 & 59.71 & 36.84 & \underline{76.47} \\
        \quad MemoryOS\incgreenl{$+$DeepSeek-R1-0528} & 62.65 & 62.79 & 78.18 & 11.83 & 77.78 & 70.50 & 78.95 & \textbf{82.35} \\
        \rowcolor{lightblue}
        \quad PersonaTree-ALL\incgreenl{$+$DeepSeek-R1-0528} & \underline{74.87} & 80.62 & 69.09 & \underline{27.96} & \underline{94.95} & 82.01 & \textbf{89.47} & \textbf{82.35} \\
        \rowcolor{lightblue}
        \quad PersonaTree-ALL\incgreenl{$+$Qwen2.5-7B-RL} & 74.70 & \underline{79.84} & \underline{80.00} & \underline{27.96} & \underline{94.95} & 82.01 & 84.21 & 64.71  \\
        \rowcolor{lightblue}
        \quad PersonaTree-ALL\incgreenl{$+$Qwen3-8B-RL} & \textbf{76.06} & \textbf{80.62} & \textbf{81.82} & \textbf{29.03} & 92.93 & \textbf{84.17} & \underline{87.72} & \underline{76.47}  \\
        \addlinespace[1pt]
        \rowcolor[rgb]{0.94,0.94,0.94} \quad \qquad w/  PersonaTree$+$Router & 74.19 & \textbf{80.62} & 67.27 & 25.81 & \textbf{95.96} & \underline{83.45} & 84.21 & \underline{76.47} \\
         \rowcolor[rgb]{0.94,0.94,0.94} \quad \qquad w/ PersonaTree & 65.70 & 71.32 & 65.45 & 18.28 & 91.92 & 69.78 & 75.44 & 64.71 \\
        \bottomrule
        \end{tabular}
    \end{threeparttable}
    }
    \caption{Main experimental results are presented on three different response models. Recall-Facts, Pref-Rec, New-Ideas, Recall-Reason, Pref-Evol, Gen-New, and Recall-User respectively denote recalling user-shared facts, providing preference-aligned recommendations, suggesting new ideas, recalling reasons behind preference updates, tracking preference evolution, generalizing to new scenarios, and recalling user-mentioned facts. \textbf{Bold} and \underline{underlined} numbers denote the best and second-best results, respectively.}
    \label{tab:main results}
\end{table*}
As shown in Table~\ref{tab:main results}, we conduct a unified evaluation across three response models: DeepSeek-V3.1, Longcat-Flash-Chat, and DeepSeek-R1-0528, while fixing DeepSeek-R1-0528 as the memory extractor for all settings. 

On DeepSeek-V3.1, our best configuration achieves $\mathrm{Overall}=71.31$, improving by $+18.68$ over Only LLM and by $+7.47$ over ALL Dialogue, and further exceeding the strongest comparative memory system, MemoryOS, by $+8.83$. On Longcat-Flash-Chat, PersonaTree-ALL with Qwen3-8B-RL attains the best overall performance with $\mathrm{Overall}=75.38$, improving by $+13.58$ over ALL Dialogue and by $+10.35$ over MemoryOS. Under this setting, all sub-metrics exhibit consistent improvements, indicating that after dialogue compression and multi-round retrieval fusion, the model can more accurately recover the user's factual background, characterize the temporal evolution of preferences, and generate new content with broader coverage.

When using DeepSeek-R1-0528 as the response model, we observe similarly substantial improvements: PersonaTree-ALL with Qwen3-8B-RL reaches $\mathrm{Overall}=76.06$, outperforming ALL Dialogue by $+11.20$ and surpassing MemoryOS by $+13.41$. In terms of fine-grained metrics, Pref-Rec and New-Ideas improve by $+18.18$ and $+17.20$ over ALL Dialogue, respectively, indicating that structured memory combined with process-aligned generation can substantially strengthen preference consistency and creative completion. Meanwhile, the method maintains stable advantages on Recall-Reason and Pref-Evo, suggesting that PersonaTree offers greater interpretability and controllability in preserving and invoking causal chains and evolution cues. More comprehensive comparative experiments, ablation analyses and visual presentations are shown in Appendix \ref{appendix Extra Experiments}.

\begin{table}[ht!]
    \centering
    \resizebox{0.7\linewidth}{!}{
    \begin{threeparttable}
        \begin{tabular}{lcccc}
        \toprule
        \multirow{2}{*}{\textbf{Methods}} & \multicolumn{3}{c}{\textbf{Evaluation Models}} & \multirow{2}{*}{\textbf{\makecell{Avg. Length \\ of Context}}} \\ 
        \addlinespace[1pt]
        \cline{2-4}
        \addlinespace[1pt]
        & \textbf{DS-V3.1} & \textbf{Longcat} & \textbf{DS-R1-0528} &  \\
        \midrule
        Only LLM & 52.63 & 54.33 & 44.14 & 0 \\
        ALL Dialogue & \textbf{63.84} & 61.80 & \underline{64.86} & 32K \\
        \addlinespace[1pt]
        \cdashline{1-5}
        \addlinespace[1pt]
        \rowcolor{lightgray}
        \multicolumn{5}{l}{\texttt{\incblue{PersonaTree}}} \\
        \rowcolor{lightgray}
        \quad $+$Qwen2.5-7B-Instruct & 55.18 & 55.18 & 50.08 & 1852.08 \\
        \rowcolor{lightgray}
        \quad $+$Qwen3-8B & 52.97 & 54.33 & 47.71 & 1392.49 \\
        \rowcolor{lightgray}
        \quad $+$GPT-4o-mini & 55.86 & 58.23 & 55.18 & 1154.35 \\
        \rowcolor{lightgray}
        \quad $+$Longcat-Flash-Chat & 58.91 & 61.63 & 60.78 & 2305.46 \\
        \rowcolor{lightgray}
        \quad $+$DeepSeek-V3.1 & 60.03 & 62.31 & 61.80 & 2227.78 \\
        \rowcolor{lightgray}
        \quad $+$DeepSeek-V3.2 & 60.32 & 63.50 & 63.16 & 2292.54 \\
        \rowcolor{lightgray}
        \quad $+$DeepSeek-R1-0528 & 60.61 & 63.33 & 63.50 & 1844.19 \\
        \rowcolor{lightgray}
        \quad $+$Gemini-3-Pro & 61.29 & 63.16 & 63.16 & 2252.89 \\
        \addlinespace[1pt]
        \cdashline{1-5}
        \addlinespace[1pt]
        \rowcolor{lightblue}
        \rowcolor{lightblue}
        \multicolumn{5}{l}{\quad \texttt{\incgreen{$+$Qwen2.5-7B-Instruct}}} \\
        \rowcolor{lightblue}
        \quad \quad $+$SFT & 60.27 & 62.82 & 60.61 & 2158.03 \\
        \rowcolor{lightblue}
        \quad \quad $+$SFT+RL & \underline{62.82} & \underline{64.35} & 64.01 & 2626.05 \\
        \rowcolor{lightblue}
        \multicolumn{5}{l}{\quad \texttt{\incgreen{$+$Qwen3-8B}}} \\
        \rowcolor{lightblue}
        \quad \quad $+$SFT & 61.29 & 62.99 & 63.33 & 2204.57 \\
        \rowcolor{lightblue}
        \quad \quad $+$SFT+RL & 61.97 & \textbf{65.20} & \textbf{65.70} & 2348.49 \\
        \bottomrule
        \end{tabular}
    \end{threeparttable}
    }
    \caption{Performance comparison of different models in generating PersonaTree memory operations. DS-V3.1, Longcat and DS-R1-0528 respectively denote DeepSeek-V3.1, Longcat-Flash-Chat and DeepSeek-R1-0528.}
    \label{tab:extraction llm}
\end{table}

\subsection{Ablation Study}
\label{main Ablation Study}
\paragraph{Effectiveness of Components.}
As presented in Table~\ref{tab:main results}, under the Qwen3-8B-RL setting we conduct a component-level ablation of the adaptive generation pipeline by comparing three inference routes: using only the lightweight fast mode (w/ PersonaTree), adding router-based triggering on top of the fast mode (w/ PersonaTree+Router), and the full agentic recall and fusion (PersonaTree-ALL). The results consistently indicate that PersonaTree alone yields robust gains, but agentic recall and fusion is critical for achieving the best performance, while the routing mechanism can closely match the full-mode performance while substantially reducing additional retrieval overhead. These gains suggest that directly injecting structured memory already covers a large portion of stable attribute-related requirements, whereas routing and agentic recall further strengthen fine-grained characterization of complex intents and evidence completion.

\paragraph{Enhancement through Training.}
As shown in Table~\ref{tab:extraction llm}, we further validate the effectiveness of training the memory-operation model. We construct two datasets, PersonaMem 15K and HaluMem 13K. For RL, we use HaluMem 13K for SFT warm-up, and additionally employ 0.5K PersonaMem for process-reward training. For the pure SFT setting, we train on PersonaMem 15K. The results show that training can substantially improve the usability of the memory encoded in PersonaTree as well as downstream reasoning quality. More importantly, whereas ALL Dialogue requires approximately 32K context, the trained PersonaTree introduces only about 2.2K--2.6K tokens of memory context on average, yet can match or exceed the ALL Dialogue baseline on Longcat-Flash-Chat and DeepSeek-R1-0528.

\paragraph{Selection of Generation Strategies.}
\begin{table}[ht!]
    \centering
    \resizebox{0.7\linewidth}{!}{
    \begin{threeparttable}
        \begin{tabular}{lcccc}
        \toprule
        &  & \multicolumn{3}{c}{\textbf{Evaluation Models}} \\
        \addlinespace[1pt]
        \cline{3-5}
        \addlinespace[1pt]
        \multirow{-2}{*}{\textbf{Extraction Models}} & \multirow{-2}{*}{\textbf{Direct}} & \textbf{DS-V3.1} & \textbf{Longcat} & \textbf{DS-R1-0528} \\
        \midrule
        \multirow{2}{*}{DeepSeek-V3.1} & No &  58.91 & 60.78 & 58.74 \\
         & Yes & \cellcolor{lightgray}\textbf{60.03\incred{+1.12}} & \cellcolor{lightgray}\textbf{62.31\incred{+1.53}} & \cellcolor{lightgray}\textbf{61.80\incred{+3.06}} \\
        \addlinespace[1pt]
        \cdashline{1-5}
        \addlinespace[1pt]
        \multirow{2}{*}{Longcat-Flash-Chat} & No & 58.57 & 59.59 & 58.40 \\
         & Yes & \cellcolor{lightgray}\textbf{58.91\incred{+0.34}} & \cellcolor{lightgray}\textbf{61.63\incred{+2.04}} & \cellcolor{lightgray}\textbf{60.78\incred{+2.38}} \\
        \addlinespace[1pt]
        \cdashline{1-5}
        \addlinespace[1pt]
        \multirow{2}{*}{DeepSeek-R1-0528} & No & 59.76 & 60.95 & 62.31 \\
         & Yes & \cellcolor{lightgray}\textbf{60.61\incred{+0.85}} & \cellcolor{lightgray}\textbf{63.33\incred{+2.38}} & \cellcolor{lightgray}\textbf{63.50\incred{+1.19}} \\
        \bottomrule
        \end{tabular}
    \end{threeparttable}
    }
    \caption{Performance comparison between direct Generation and extract-then-transform approaches for generating PersonaTree memory operations.}
    \label{tab:direct extraction}
\end{table}
We compare two strategies for generating PersonaTree operations: direct generation (the model directly outputs a tree-operation sequence conditioned on the dialogue) and extract-then-transform (first extracting personalized information from the dialogue and then mapping it into tree operations). As shown in Table~\ref{tab:direct extraction}, direct generation achieves consistent advantages in all combinations, with improvements ranging from approximately $+0.34$ to $+3.06$. The two-stage strategy accumulates errors in the intermediate representation, often losing fine-grained semantic and temporal cues needed for accurate tree operations. Based on this finding, we adopt direct generation as the default PersonaTree operation-generation approach in the remainder of this paper.

\subsection{Hyperparameter Analysis}
\begin{table}[ht!]
    \centering
    \resizebox{0.7\linewidth}{!}{
    \begin{threeparttable}
        \begin{tabular}{lccccccc}
        \toprule
        \multirow{2}{*}{\textbf{Evaluation Models}} & \multicolumn{7}{c}{\textbf{Chunk Window Size}} \\
        \addlinespace[1pt]
        \cline{2-8}
        \addlinespace[2pt]
         & \textbf{1} & \textbf{3} & \textbf{5} & \textbf{7} & \textbf{10} & \textbf{13} & \textbf{15} \\
        \midrule
        \multicolumn{8}{l}{\texttt{\incblue{DeepSeek-R1-0528 (Tree Extraction)}}} \\
        \quad DeepSeek-V3.1 & 59.42 & 60.61 & 58.74 & 58.57 & 60.10 & 59.25 & 58.74 \\
        \quad Longcat-Flash-Chat & 62.82 & 63.33 & 62.48 & 62.65 & 62.65 & 62.31 & 62.48 \\
        \quad DeepSeek-R1-0528 & 60.95 & 63.50 & 59.93 & 60.27 & 59.25 & 62.65 & 61.29 \\
        \addlinespace[1pt]
        \cdashline{1-8}
        \addlinespace[1pt]
        \multicolumn{8}{l}{\texttt{\incblue{Qwen2.5-7B-RL (Tree Extraction)}}} \\
        \quad DeepSeek-V3.1 & 60.78 & 62.82 & 58.91 & 61.12 & 59.25 & 57.56 & 59.59 \\
        \quad Longcat-Flash-Chat & 62.14 & 64.35 & 63.67 & 61.29 & 64.18 & 62.82 & 61.97 \\
        \quad DeepSeek-R1-0528 & 62.82 & 64.01 & 60.78 & 62.82 & 60.95 & 61.29 & 62.14 \\
        \bottomrule
        \end{tabular}
    \end{threeparttable}
    }
    \caption{Sensitivity analysis on the size of segmented dialogue chunks.}
    \label{tab:chunk size}
\end{table}

We analyze the impact of the dialogue chunking window on the quality of tree-operation generation, where each dialogue chunk consists of $w$ consecutive dialogue turns. As shown in Table~\ref{tab:chunk size}, regardless of whether DeepSeek-R1-0528 or Qwen2.5-7B-RL is used as the tree-operation generation model, $w=3$ yields the most stable and overall best performance across all three evaluation models. When the window is too small ($w=1$), chunking becomes overly fragmented and tends to introduce noisy writes; when the window is too large ($w\ge 10$), the within-chunk information density increases substantially, with more frequent cross-topic mixing and timeline collapsing, making critical cues more likely to be diluted or missed. Accordingly, we adopt $w=3$ for dialogue chunking and tree-operation generation in our experiments.

\section{Conclusion}
This paper studies memory evolution for long-term personalized dialogue in memory systems and proposes the \textbf{Inside Out} framework, which uses an explicit, structured \textbf{PersonaTree} as long-term memory to maintain personalized states under unbounded interactions. Concretely, we build a hierarchical schema grounded in the biopsychosocial model and develop an iterative tree-update mechanism. We then train a lightweight \textbf{MemListener} with process-reward RL to compress unstructured dialogue streams into executable tree operations. At inference time, we design an adaptive generation pipeline. Experiments show that PersonaTree-driven personalization consistently outperforms existing baselines across multiple response models, and further highlight the potential of using small models for memory maintenance.

\section*{Limitations}
This work focuses on the structured evolution of long-term personalized memory, and the current implementation and empirical validation delineate clear directions for future extension:

\textbf{Scope of the schema and PersonaTree.} 
We adopt a hierarchical schema grounded in the biopsychosocial model to define the writable space and capacity constraints, yielding a consistent and controllable representation of long-term memory. For finer-grained domain knowledge, task-skill profiles, or cross-domain user states, the schema can be further extended into composable subtrees or plugin-style modules to accommodate broader application needs.

\textbf{Applicability of the memory-evolution strategy.} 
PersonaTree is iteratively updated via atomic tree operations, with add/modify/delete semantics uniformly expressed as trunk and leaf-level text rewrites. This abstraction is effective for preserving stable core traits and compressible summaries; for memory forms requiring stronger temporal constraints, evidence provenance, or multi-version coexistence, additional metadata such as explicit timestamps, confidence scores, and source pointers can be incorporated to improve traceability and controllability.

\textbf{Engineering extensions for privacy and governance.}
As an explicit long-term memory carrier, PersonaTree naturally supports access control, interpretable edits, and revocability. For real-world deployment, it can be further complemented with user-facing memory management, sensitivity-aware field stratification, and data-minimization storage policies to meet stricter governance requirements.

\bibliographystyle{plainnat}
\bibliography{paper}

\clearpage

\appendix

\section{Rethinking Personalization}
In Human-AI interaction, building agents capable of deeply personalized dialogue has long been a central goal. However, the dominant research paradigm largely concentrates on personalization via explicit profiles \citep{li2016persona,zhang2018personalizing}. Under this setting, researchers typically provide either a structured or unstructured persona description, or a set of persona-related texts to be retrieved, and the model is tasked with generating user-aligned responses conditioned on this static, pre-specified information \citep{huang2023learning,li2025persona}.

Although this paradigm offers advantages in controllability and evaluation convenience, it deviates substantially from real-world interpersonal interaction. This deviation is mainly reflected in:
\begin{itemize}
    \item \textbf{Misalignment of information sources:} In everyday life, our understanding of a person's traits rarely comes from a self-introduction document; instead, it is implicitly and dynamically constructed from long-term interaction history \citep{han2023personapkt,otsuka2024user}.
    \item \textbf{Limited characterization of persona:} Explicit persona descriptions are often highly abstracted and simplified, failing to capture subtle linguistic styles, background knowledge, distinctive interaction patterns, and affective dynamics that emerge in authentic conversations \citep{chen2025deeper}.
    \item \textbf{A static assumption of persona:} Real personality traits and linguistic styles vary across contexts and conversations, whereas static-profile approaches struggle to model such adaptive dynamics \citep{ong2025towards}.
\end{itemize}

Given these limitations, our study targets a more challenging and more realistic core problem:
\textit{How can a model, relying solely on a long personalized dialogue history, learn and emulate one participant's implicit persona to generate responses that remain consistent in style, content, and relational stance?}

Our motivation is to help bridge this gap, with three primary implications:
\begin{itemize}
    \item \textbf{Improving the realism of personalized dialogue:} enabling a shift from role-playing to faithful imitation, producing responses that are more natural, credible, and person-like.
    \item \textbf{Advancing deep personalization modeling:} moving beyond understanding facts about a person toward modeling how a person becomes who they are.
    \item \textbf{Expanding real-world applicability:} in emerging applications such as personalized assistants and affective companions, the ability to reproduce individual styles from historical data is crucial.
\end{itemize}

\section{MemListener Training}
\label{MemListener Training}
\begin{table}[ht!]
    \centering
    \begin{tabular}{lc}
    \toprule
    \textbf{Parameter} & \textbf{Value} \\
    \midrule
    Training type & Full fine-tuning \\
    Model precision & bfloat16 \\
    Learning rate & $1\times10^{-6}$ \\
    Per Device Train Batch Size & 1 \\
    Training batch size & 1 \\
    Gradient accumulation steps & 8 \\
    Number of epochs & 1 \\
    Warmup ratio & 0.01 \\
    Max gradient norm & 1.0 \\
    Max sequence length & 11264 \\
    Max generation length & 512 \\
    Number of generations & 8 \\
    Temperature & 1.0 \\
    Top-$p$ & 0.9 \\
    Top-$k$ & 50 \\
    Clipping $\epsilon$ & 0.2 / 0.28 \\
    $\beta$ (KL control) & 0.001 \\
    Dynamic sampling & Enabled \\
    Max resample times & 3 \\
    \bottomrule
    \end{tabular}
    \caption{Key hyperparameters for DAPO training.}
    \label{tab:dapo_hyperparams}
\end{table}
After the SFT warm-up, we continue alignment with the remaining data using RL driven by process-based rewards. We set the model's maximum context length to 11K tokens, with the input length capped at 10K, which constitutes a typical ultra-long sequence optimization setting. If we were to adopt sample-level group-relative policy optimization (GRPO), the key decision signals in long sequences would be easily diluted by within-sample averaging; moreover, when group-wise sampling under the same input yields outputs that are all correct or all incorrect, the advantage term degenerates, resulting in insufficient effective gradients. Consequently, training stability and sample efficiency are constrained. 

Therefore, the RL stage is conducted using the Swift RLHF framework~\citep{zhao2024swift} with the DAPO algorithm (Table~\ref{tab:dapo_hyperparams}). All model parameters are updated via full fine-tuning, and training is performed in bfloat16 precision to balance numerical stability and memory efficiency. We further employ a dynamic, LLM-as-a-judge evaluation strategy, using Qwen3-32B (reasoning mode) as the discriminator to score the gap between the model’s prediction and the ground truth. The judge is prompted using the template in Table~\ref{zztab:8}, and its assessment signal is used to guide optimization during RL. For transparency and reproducibility, we release the complete training scripts in our public repository.

A learning rate of $1\times10^{-6}$ is adopted, together with a warmup ratio of 0.01. Due to memory constraints, the per-device training batch size is set to 1, while the effective batch size is increased using gradient accumulation over 8 steps. Gradient norms are clipped to 1.0 to ensure stable optimization. Training is performed for a single epoch.

For each input prompt, 8 candidate responses are sampled with a maximum generation length of 512 tokens and a maximum context length of 11,264 tokens. Stochastic decoding is controlled using temperature = 1.0, top-$p$ = 0.9, and top-$k$ = 50. Policy updates use asymmetric clipping with $\epsilon = 0.2$ and $\epsilon_{\text{high}} = 0.28$.

A KL-control coefficient $\beta$ is introduced to regulate the divergence between the optimized policy and the reference model. Larger $\beta$ values enforce stronger regularization toward the reference policy. In our GRPO-based setting, $\beta$ is set to 0.001. Dynamic sampling is enabled to enhance response diversity, with the maximum number of resampling attempts limited to 3.

During training data construction, we select subsets from HaluMem and PersonaMem that are relevant to implicitly characterizing user-specific attributes as the raw corpus sources. Using the dynamic PersonaTree evolution procedure described in Section~\ref{Dynamic PersonaTree Evolution} as the backbone, we invoke DeepSeek-R1-0528 to generate supervision signals for training. Concretely, for each dialogue segment, we prompt the generator to produce an executable operation sequence and its corresponding post-update tree state under the given schema and update constraints, thereby mapping raw dialogues into structured samples with ground-truth operations and versioned tree evolution. To control noise and spurious correlations, we manually verify the synthesized samples and filter out those with invalid operation syntax, incorrect path references, or semantically inconsistent writes. The specific prompts are shown in Table \ref{zztab:8}.

\section{Extra Experiments}
\label{appendix Extra Experiments}
\begin{table*}[ht!]
    \centering
    \resizebox{1.0\linewidth}{!}{
    \begin{threeparttable}
        \begin{tabular}{lcccccccc}
        \toprule
        \textbf{Methods} & \textbf{Overall} & \textbf{Recall-Facts} & \textbf{Pref-Rec} & \textbf{New-Ideas} & \textbf{Recall-Reason} & \textbf{Pref-Evol} & \textbf{Gen-New} & \textbf{Recall-User} \\ 
        \midrule
        \multicolumn{9}{l}{\texttt{\incblue{DeepSeek-V3.1 (Response)}}} \\
        \quad Only LLM & 52.63 & 59.69 & 43.64 & 5.38 & 80.81 & \underline{65.47} & 42.11 & 52.94 \\
        \quad ALL Dialogue & \underline{63.84} & 78.29 & 61.82 & \underline{12.90} & 81.82 & \underline{65.47} & \underline{77.19} & \textbf{76.47} \\
        \quad LangMem\incgreenl{$+$DeepSeek-V3.1} & 56.37 & 72.87 & 52.73 & 9.68 & 78.79 & 61.15 & 50.88 & 47.06 \\
        \quad Mem0\incgreenl{$+$DeepSeek-V3.1} & 61.80 & \underline{82.95} & 47.27 & 7.53 & 83.84 & 63.31 & 71.93 & \underline{70.59} \\
        \quad A-Mem\incgreenl{$+$DeepSeek-V3.1} & 60.61 & 76.74 & 43.64 & 6.45 & \underline{85.86} & 64.03 & 71.93 & \textbf{76.47} \\
        \quad MemoryOS\incgreenl{$+$DeepSeek-V3.1} & 61.97 & 75.19 & \textbf{67.27} & 9.68 & 80.81 & 63.31 & 73.68 & \underline{70.59} \\
        \rowcolor{lightblue}
        \quad PersonaTree-ALL\incgreenl{$+$DeepSeek-V3.1} & \textbf{70.80} & \textbf{88.37} & \underline{63.64} & \textbf{24.73} & \textbf{86.87} & \textbf{68.35} & \textbf{91.23} & \underline{70.59} \\
        \addlinespace[1pt]
        \cdashline{1-9}
        \addlinespace[1pt]
        \multicolumn{9}{l}{\texttt{\incblue{Longcat-Flash-Chat (Response)}}} \\
        \quad Only LLM & 54.33 & 62.02 & 43.64 & 8.60 & 84.85 & \textbf{71.22} & 28.07 & 52.94 \\
        \quad ALL Dialogue & 61.80 & 79.07 & 63.64 & 15.05 & \underline{87.88} & 53.96 & \underline{68.42} & \underline{70.59} \\
        \quad LangMem\incgreenl{$+$DeepSeek-V3.1} & 57.39 & 75.97 & 50.91 & 10.75 & 79.80 & 63.31 & 45.61 & 52.94 \\
        \quad Mem0\incgreenl{$+$DeepSeek-V3.1} & 60.27 & 77.52 & 54.55 & 9.68 & 85.86 & 64.03 & 50.88 & \textbf{76.47} \\
        \quad A-Mem\incgreenl{$+$DeepSeek-V3.1} & 60.44 & \underline{80.62} & 50.91 & 8.60 & 85.86 & 66.19 & 47.37 & \underline{70.59} \\
        \quad MemoryOS\incgreenl{$+$DeepSeek-V3.1} & \underline{64.52} & 79.84 & \underline{65.45} & \underline{16.13} & \underline{87.88} & 64.75 & 66.67 & 64.71 \\
        \rowcolor{lightblue}
        \quad PersonaTree-ALL\incgreenl{$+$DeepSeek-V3.1} & \textbf{73.34} & \textbf{88.37} & \textbf{69.09} & \textbf{31.18} & \textbf{91.92} & \underline{69.78} & \textbf{87.72} & \textbf{76.47} \\
        \addlinespace[1pt]
        \cdashline{1-9}
        \addlinespace[1pt]
        \multicolumn{9}{l}{\texttt{\incblue{DeepSeek-R1-0528 (Response)}}} \\
        \quad Only LLM & 44.14 & 30.23 & 38.18 & 19.35 & 76.77 & 55.40 & 38.60 & 41.18 \\
        \quad ALL Dialogue & \underline{64.86} & \underline{69.77} & 63.64 & 11.83 & \underline{84.85} & \underline{73.38} & \underline{84.21} & 70.59 \\
        \quad LangMem\incgreenl{$+$DeepSeek-V3.1} & 56.54 & 62.79 & 54.55 & 18.28 & 78.79 & 66.19 & 49.12 & 41.18 \\
        \quad Mem0\incgreenl{$+$DeepSeek-V3.1} & 47.71 & 32.56 & 52.73 & 21.51 & 73.74 & 57.55 & 47.37 & 58.82 \\
        \quad A-Mem\incgreenl{$+$DeepSeek-V3.1} & 47.03 & 27.13 & 45.45 & \textbf{27.96} & 76.77 & 55.40 & 43.86 & \underline{76.47} \\
        \quad MemoryOS\incgreenl{$+$DeepSeek-V3.1} & 61.97 & 64.34 & \textbf{76.36} & 11.83 & 82.83 & 66.19 & 73.68 & \underline{76.47} \\
        \rowcolor{lightblue}
        \quad PersonaTree-ALL\incgreenl{$+$DeepSeek-V3.1} & \textbf{74.53} & \textbf{81.40} & \underline{70.91} & \underline{26.88} & \textbf{92.93} & \textbf{82.01} & \textbf{87.72} & \textbf{82.35} \\
        \bottomrule
        \end{tabular}
    \end{threeparttable}
    }
    \caption{Extra experimental results are presented on three different response models. Recall-Facts, Pref-Rec, New-Ideas, Recall-Reason, Pref-Evol, Gen-New, and Recall-User respectively denote recalling user-shared facts, providing preference-aligned recommendations, suggesting new ideas, recalling reasons behind preference updates, tracking preference evolution, generalizing to new scenarios, and recalling user-mentioned facts. \textbf{Bold} and \underline{underlined} numbers denote the best and second-best results, respectively.}
    \label{tab:extra results}
\end{table*}

\subsection{Baselines}
\label{add_Baselines}
To evaluate the effectiveness of our proposed approach, we compared it against six baseline methods. These include two standard interaction paradigms (Only LLM and Full Context) and four state-of-the-art memory management frameworks designed for LLM-based agents. All memory management frameworks were evaluated under their officially recommended best configurations.

\paragraph{Only LLM.} As a foundational baseline, we employ the LLM directly without providing any historical conversation data. In this setting, the model operates in a stateless manner, relying solely on its pre-trained parametric knowledge and internal reasoning capabilities to address user queries. This method serves as a lower bound, isolating the model's intrinsic commonsense reasoning from its ability to recall specific interactional details.

\paragraph{Full Context.} This method involves concatenating the entire chronological history of the conversation and inputting it into the LLM’s context window for every interaction. By providing the model with complete access to all prior dialogue, this approach serves as a theoretical upper bound for retrieval accuracy within the limits of the model's context window. However, it effectively ignores the challenges of memory selection and computational efficiency.

\paragraph{LangMem.} LangMem is a framework designed to enable agents to learn and adapt through continuous interactions. It provides a suite of functional primitives that allow agents to manage memory within the active conversational flow ("hot path") while utilizing a background manager to asynchronously extract, consolidate, and update knowledge. LangMem integrates natively with the LangGraph ecosystem, offering a core memory API that supports prompt refinement and long-term consistency across sessions. By separating immediate memory management tools from background consolidation processes, it aims to maintain consistent agent behavior without increasing latency during inference.

\paragraph{Mem0~\citep{chhikara2025mem0}.} Mem0 addresses the limitations of fixed context windows by introducing a scalable, memory-centric architecture. It employs a multi-level memory structure that retains User, Session, and Agent states to facilitate adaptive personalization. A key feature of Mem0 is its utilization of graph-based memory representations to capture complex relational structures between conversational elements. This approach allows for the dynamic extraction and retrieval of salient information, optimizing for both latency and token cost. Mem0 is designed to be production-ready, focusing on reducing the computational overhead typically associated with full-context processing while maintaining high retrieval accuracy in multi-session dialogues.

\paragraph{A-Mem (Agentic Memory)~\citep{xu2025mem}.} A-Mem proposes a self-organizing memory system inspired by the Zettelkasten knowledge management method. Unlike traditional static storage, A-Mem enables agents to dynamically organize memories through intelligent indexing and linking. When new information is ingested, the system generates comprehensive notes with structured attributes—such as contextual descriptions and tags—and establishes connections with historical data. A distinctive feature of A-Mem is its support for "memory evolution," where the integration of new experiences can trigger updates to the representations of existing memories. This agent-driven mechanism allows the memory network to continuously refine its structure and understanding over time.

\paragraph{MemoryOS~\citep{kang2025memory}.} Drawing inspiration from operating system principles, MemoryOS introduces a hierarchical storage architecture designed to manage agent memory comprehensively. The system comprises four core modules: Storage, Updating, Retrieval, and Generation. It organizes memory into three distinct levels: short-term, mid-term, and long-term personal memory. To manage data flow between these levels, MemoryOS employs specific strategies such as a dialogue-chain-based First-In-First-Out (FIFO) principle for short-to-mid-term updates and a segmented page organization strategy for mid-to-long-term consolidation. This hierarchical approach aims to maximize context coherence and personalization by mimicking the efficient resource management found in traditional operating systems.

\subsection{Extra Experiments Results}
To validate the robustness of our approach for the tree-operation generator, we further adopt DeepSeek-V3.1 as a unified extraction model to conduct comparative evaluations across all methods, and we report results separately under three different response models (DeepSeek-V3.1, Longcat-Flash-Chat, and DeepSeek-R1-0528; see Table~\ref{tab:extra results}) to minimize potential biases and hallucination effects introduced by any particular generation component. The results show that, even when replacing the extraction model DeepSeek-R1-0528 with DeepSeek-V3.1, our method (PersonaTree-ALL) still achieves the best Overall scores under all three response models, reaching 70.80, 73.34, and 74.53, respectively. This corresponds to improvements of +18.17/+19.01/+30.39 over Only LLM and +6.96/+11.54/+9.67 over ALL Dialogue, and it substantially outperforms the representative memory baseline MemoryOS by +8.83/+8.82/+12.56. From a metric-wise perspective, the gains are particularly pronounced on dimensions requiring stronger detail completion and open-ended generation; for example, New-Ideas improves over ALL Dialogue by +11.83 (24.73 vs.\ 12.90), +16.13 (31.18 vs.\ 15.05), and +15.05 (26.88 vs.\ 11.83), respectively. Meanwhile, Recall-Facts, Pref-Rec, and Gen-New also exhibit consistent improvements. These findings indicate that PersonaTree's structured memory representation and retrieval-augmented generation mechanism do not rely on any specific extraction model; rather, the benefits transfer stably across extractors and response models, further supporting the generality and effectiveness of the proposed method.

\paragraph{Ablation Study.}
Beyond the partial ablation studies reported in Table~\ref{tab:main results}, we further conduct a more comprehensive ablation analysis on the remaining configurations in Tables~\ref{tab:main results} and~\ref{tab:extra results}, with detailed results presented in Tables~\ref{tab:ablation results} and~\ref{tab:ablation results 2}. Overall, these additional ablation findings are consistent with the observations in Table~\ref{tab:main results}, further corroborating the effectiveness and contributions of the key components across different experimental settings, and providing stronger empirical support for the main conclusions in Section~\ref{main Ablation Study}.

\subsection{Visualization}
\begin{figure*}[ht!]
    \centering
    \includegraphics[width=\textwidth]{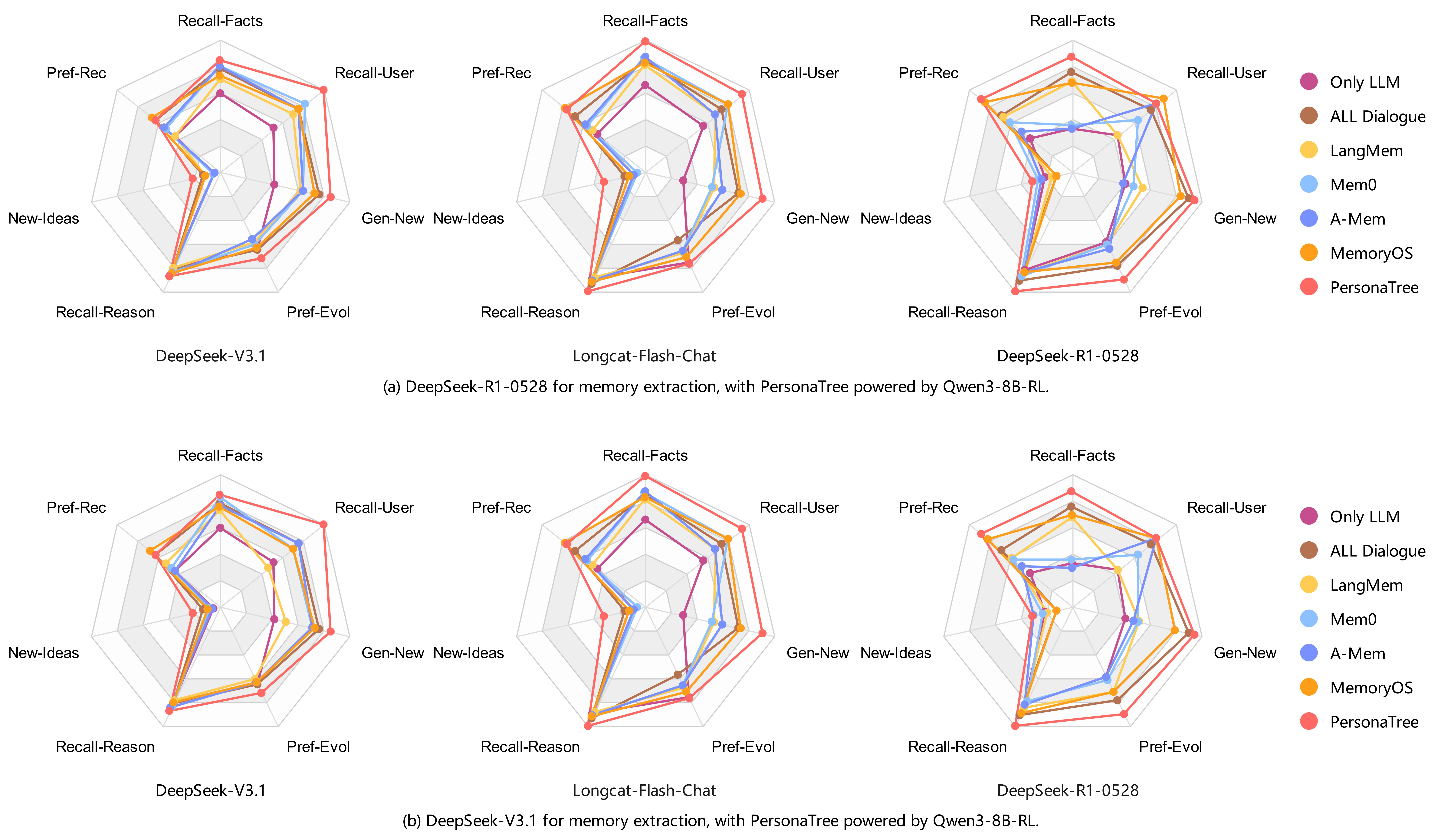}
    \caption{Radar-chart comparison of PersonaTree and baselines across multi-dimensional capability metrics under two memory-extraction settings (DeepSeek-R1-0528, DeepSeek-V3.1) and three response models (DeepSeek-V3.1, Longcat-Flash-Chat, DeepSeek-R1-0528)}
    \label{fig:leidatu}
\end{figure*}

To more intuitively illustrate the differences among the methods in Tables~\ref{tab:main results} and~\ref{tab:extra results} across diverse capability dimensions, we further provide radar-chart visualizations (see Figure~\ref{fig:leidatu}), corresponding to two memory-extraction settings: (a) using DeepSeek-R1-0528 as the extractor; and (b) using DeepSeek-V3.1 as the extractor. For each setting, we conduct a unified comparison under three response models: DeepSeek-V3.1, Longcat-Flash-Chat, and DeepSeek-R1-0528. Overall, PersonaTree exhibits a more outward-expanded polygonal profile across both extraction settings and all three response models, indicating that its gains are not concentrated on a single metric but instead span multiple dimensions, including factual recall, preference consistency, preference evolution, and new-content generation.

Moreover, the trends across the two subplots are highly consistent, suggesting that PersonaTree's improvements are robust to the choice of memory-extraction model: even when the extractor is replaced, the method maintains stable advantages across the multi-dimensional metrics.

\begin{figure*}[h!]
    \centering
    \includegraphics[width=0.8\textwidth]{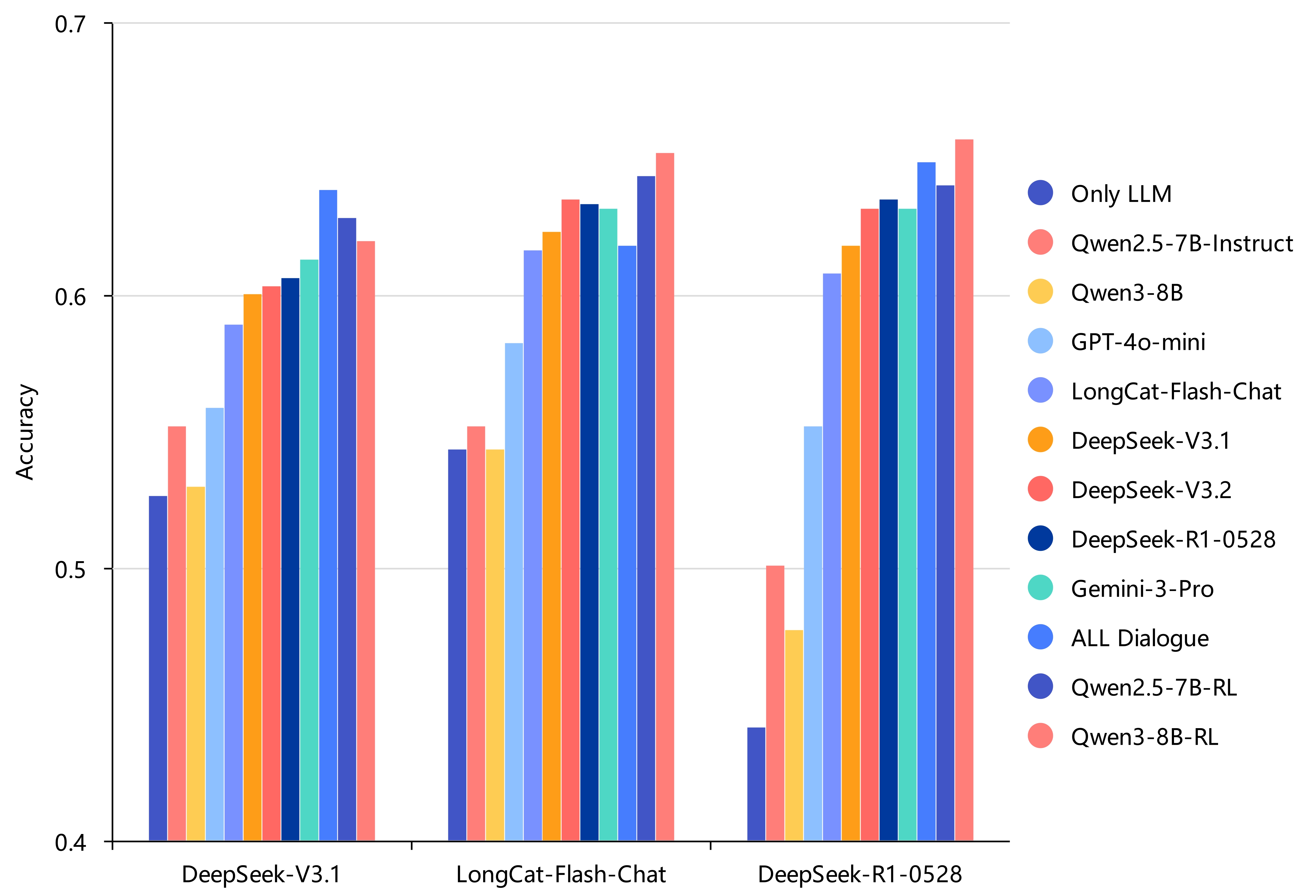}
    \caption{Overall performance of memory-operation models across three response models.}
    \label{fig:zhuzhuang}
\end{figure*}

On the other hand, Figure~\ref{fig:zhuzhuang} reports the overall performance of different memory-operation models under three response models. We observe that untrained extractors generally lag behind trained counterparts, while the two-stage training paradigm (SFT+RL) yields stable and transferable improvements. Meanwhile, compared with ALL Dialogue, which requires substantially longer context, the trained PersonaTree achieves higher accuracy with only a relatively short memory context, further highlighting the advantages of structured memory in terms of information compression and utilization efficiency.

\begin{figure*}[h!]
    \centering
    \includegraphics[width=0.8\textwidth]{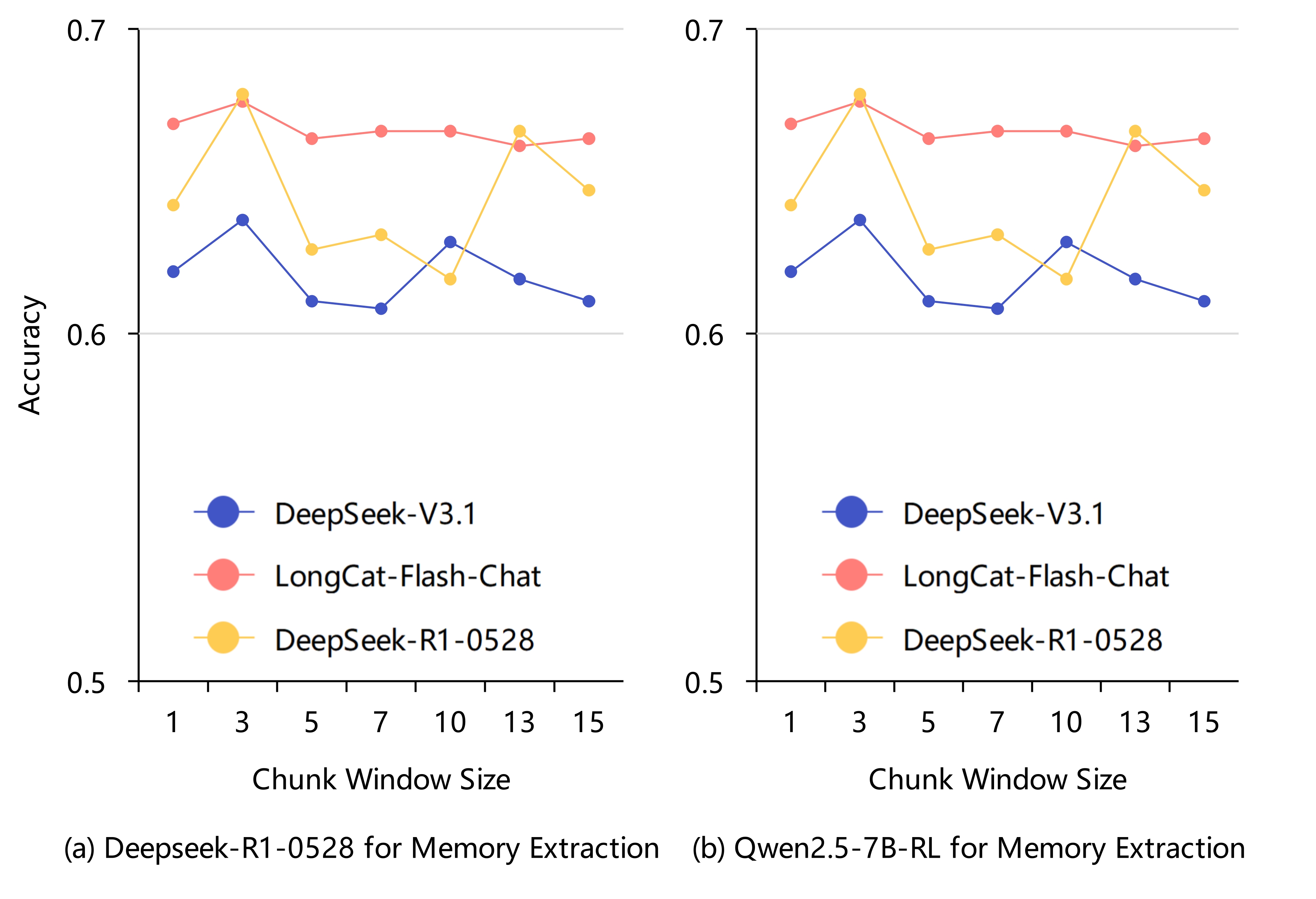}
    \caption{Effect of dialogue chunking window size on performance across three response models.}
    \label{fig:zhexian}
\end{figure*}

Figure~\ref{fig:zhexian} characterizes how the dialogue chunking window size $w$ influences performance. Across settings, the curves consistently exhibit the pattern that a moderate window is optimal. In particular, values around $w=3$ are more stable and overall superior under all three response models, suggesting that this configuration strikes a more appropriate trade-off between contextual sufficiency and update frequency.

\begin{table*}[ht!]
    \centering
    \resizebox{1.0\linewidth}{!}{
    \begin{threeparttable}
        \begin{tabular}{lcccccccc}
        \toprule
        \textbf{Methods} & \textbf{Overall} & \textbf{Recall-Facts} & \textbf{Pref-Rec} & \textbf{New-Ideas} & \textbf{Recall-Reason} & \textbf{Pref-Evol} & \textbf{Gen-New} & \textbf{Recall-User} \\ 
        \midrule
        \multicolumn{9}{l}{\texttt{\incblue{DeepSeek-R1-0528 (Tree Extraction)}}} \\
        \addlinespace[1pt]
        \cdashline{1-9}
        \addlinespace[1pt]
        \multicolumn{9}{l}{\texttt{\incgreen{$+$DeepSeek-V3.1 (Response)}}} \\
        \quad PersonaTree-ALL & \textbf{71.14} & \textbf{88.37} & 56.36 & \textbf{22.58} & \textbf{88.89} & \textbf{71.22} & \textbf{89.47} & \textbf{88.24} \\
        \quad PersonaTree$+$Router & 69.61 & 87.60 & \textbf{60.00} & 17.20 & \textbf{88.89} & 69.06 & 85.96 & \textbf{88.24} \\
        \quad Only PersonaTree & 60.61 & 77.52 & 49.09 & 6.45 & 78.79 & 66.19 & 73.68 & 70.59 \\
        \addlinespace[1pt]
        \cdashline{1-9}
        \addlinespace[1pt]
        \multicolumn{9}{l}{\texttt{\incgreen{$+$Longcat-Flash-Chat (Response)}}} \\
        \quad PersonaTree-ALL & \textbf{72.67} & 88.37 & \textbf{72.73} & \textbf{26.88} & \textbf{92.93} & 69.78 & \textbf{85.96} & 64.71 \\
        \quad PersonaTree$+$Router & 70.46 & \textbf{89.15} & 60.00 & 19.35 & 87.88 & \textbf{71.94} & 80.70 & \textbf{94.12} \\
        \quad Only PersonaTree & 63.33 & 75.97 & 60.00 & 11.83 & 88.89 & 66.91 & 70.18 & 58.82 \\
        \addlinespace[1pt]
        \cdashline{1-9}
        \addlinespace[1pt]
        \multicolumn{9}{l}{\texttt{\incgreen{$+$DeepSeek-R1-0528 (Response)}}} \\
        \quad PersonaTree-ALL & \textbf{74.87} & 80.62 & 69.09 & \textbf{27.96} & \textbf{94.95} & 82.01 & \textbf{89.47} & \textbf{82.35} \\
        \quad PersonaTree$+$Router & 73.68 & \textbf{82.17} & \textbf{74.55} & 23.66 & 92.93 & \textbf{83.45} & 84.21 & 52.94 \\
        \quad Only PersonaTree & 63.50 & 67.44 & 63.64 & 16.13 & 88.89 & 69.06 & 75.44 & 58.82 \\
        \midrule
        \multicolumn{9}{l}{\texttt{\incblue{Qwen2.5-7B-rl (Tree Extraction)}}} \\
        \addlinespace[1pt]
        \cdashline{1-9}
        \addlinespace[1pt]
        \multicolumn{9}{l}{\texttt{\incgreen{$+$DeepSeek-V3.1 (Response)}}} \\
        \quad PersonaTree-ALL & \textbf{71.31} & \textbf{89.15} & \textbf{61.82} & 23.66 & \textbf{91.92} & 69.06 & \textbf{85.96} & \textbf{76.47} \\
        \quad PersonaTree$+$Router & 70.80 & 87.60 & 60.00 & \textbf{25.81} & 87.88 & \textbf{71.22} & \textbf{85.96} & 70.59 \\
        \quad Only PersonaTree & 62.82 & 77.52 & \textbf{61.82} & 8.60 & 89.90 & 63.31 & 70.18 & 64.71 \\
        \addlinespace[1pt]
        \cdashline{1-9}
        \addlinespace[1pt]
        \multicolumn{9}{l}{\texttt{\incgreen{$+$Longcat-Flash-Chat (Response)}}} \\
        \quad PersonaTree-ALL & \textbf{73.34} & \textbf{93.02} & \textbf{67.27} & \textbf{27.96} & \textbf{92.93} & 67.63 & \textbf{89.47} & 70.59 \\
        \quad PersonaTree$+$Router & 72.33 & 92.25 & \textbf{67.27} & 23.66 & 91.92 & \textbf{68.35} & 85.96 & \textbf{76.47} \\
        \quad Only PersonaTree & 64.35 & 79.07 & 63.64 & 11.83 & 88.89 & 65.47 & 70.18 & 70.59 \\
        \addlinespace[1pt]
        \cdashline{1-9}
        \addlinespace[1pt]
        \multicolumn{9}{l}{\texttt{\incgreen{$+$DeepSeek-R1-0528 (Response)}}} \\
        \quad PersonaTree-ALL & \textbf{74.70} & 79.84 & \textbf{80.00} & 27.96 & \textbf{94.95} & \textbf{82.01} & \textbf{84.21} & 64.71 \\
        \quad PersonaTree$+$Router & 74.19 & \textbf{80.62} & 76.36 & \textbf{30.11} & 93.94 & 80.58 & \textbf{84.21} & 58.82 \\
        \quad Only PersonaTree & 64.01 & 68.99 & 63.64 & 12.90 & 84.85 & 71.22 & 80.70 & \textbf{70.59} \\
        \bottomrule
        \end{tabular}
    \end{threeparttable}
    }
    \caption{Ablation study of PersonaTree components. Recall-Facts, Pref-Rec, New-Ideas, Recall-Reason, Pref-Evol, Gen-New, and Recall-User respectively denote recalling user-shared facts, providing preference-aligned recommendations, suggesting new ideas, recalling reasons behind preference updates, tracking preference evolution, generalizing to new scenarios, and recalling user-mentioned facts. \textbf{Bold} numbers indicate the best results within each group.}
    \label{tab:ablation results}
\end{table*}
\begin{table*}[ht!]
    \centering
    \resizebox{1.0\linewidth}{!}{
    \begin{threeparttable}
        \begin{tabular}{lcccccccc}
        \toprule
        \textbf{Methods} & \textbf{Overall} & \textbf{Recall-Facts} & \textbf{Pref-Rec} & \textbf{New-Ideas} & \textbf{Recall-Reason} & \textbf{Pref-Evol} & \textbf{Gen-New} & \textbf{Recall-User} \\ 
        \midrule
        \multicolumn{9}{l}{\texttt{\incblue{DeepSeek-V3.1 (Tree Extraction)}}} \\
        \addlinespace[1pt]
        \cdashline{1-9}
        \addlinespace[1pt]
        \multicolumn{9}{l}{\texttt{\incgreen{$+$DeepSeek-V3.1 (Response)}}} \\
        \quad PersonaTree-ALL & \textbf{70.80} & \textbf{88.37} & \textbf{63.64} & \textbf{24.73} & 86.87 & 68.35 & \textbf{91.23} & \textbf{70.59} \\
        \quad PersonaTree$+$Router & 69.61 & 85.27 & 61.82 & 18.28 & \textbf{89.90} & \textbf{70.50} & 87.72 & \textbf{70.59} \\
        \quad Only PersonaTree & 60.03 & 74.02 & 56.60 & 6.59 & 82.47 & 62.50 & 71.93 & 64.71 \\
        \addlinespace[1pt]
        \cdashline{1-9}
        \addlinespace[1pt]
        \multicolumn{9}{l}{\texttt{\incgreen{$+$Longcat-Flash-Chat (Response)}}} \\
        \quad PersonaTree-ALL & \textbf{73.34} & 88.37 & \textbf{69.09} & \textbf{31.18} & \textbf{91.92} & \textbf{69.78} & \textbf{87.72} & \textbf{76.47} \\
        \quad PersonaTree$+$Router & 70.63 & \textbf{91.47} & 63.64 & 21.51 & 89.90 & 67.63 & 84.21 & 70.59 \\
        \quad Only PersonaTree & 62.31 & 78.29 & 54.55 & 10.75 & 86.87 & 66.91 & 63.16 & 64.71 \\
        \addlinespace[1pt]
        \cdashline{1-9}
        \addlinespace[1pt]
        \multicolumn{9}{l}{\texttt{\incgreen{$+$DeepSeek-R1-0528 (Response)}}} \\
        \quad PersonaTree-ALL & \textbf{74.53} & \textbf{81.40} & 70.91 & \textbf{26.88} & \textbf{92.93} & \textbf{82.01} & \textbf{87.72} & \textbf{82.35} \\
        \quad PersonaTree$+$Router & 73.51 & \textbf{81.40} & \textbf{76.36} & 24.73 & \textbf{92.93} & 79.14 & 84.21 & 76.47 \\
        \quad Only PersonaTree & 61.80 & 68.22 & 52.73 & 13.98 & 82.83 & 71.94 & 73.68 & 58.82 \\
        \bottomrule
        \end{tabular}
    \end{threeparttable}
    }
    \caption{Ablation Study II of PersonaTree Components. Recall-Facts, Pref-Rec, New-Ideas, Recall-Reason, Pref-Evol, Gen-New, and Recall-User respectively denote recalling user-shared facts, providing preference-aligned recommendations, suggesting new ideas, recalling reasons behind preference updates, tracking preference evolution, generalizing to new scenarios, and recalling user-mentioned facts. \textbf{Bold} numbers indicate the best results within each group.}
    \label{tab:ablation results 2}
\end{table*}

\section{Initial PersonaTree Instance}
\label{Initial PersonaTree Instance}
To delineate the theoretical boundaries of the memory tree, we construct a hierarchical Schema based on the Biopsychosocial model, scientifically decomposing user characteristics into three core dimensions: (1) Biological Aspects: Establishes biological traits by referencing theories in human morphology, human physiology, chronobiology, etc. (2) Psychological Aspects: Deeply mines psychological and characterological traits through cognitive psychology, the Big Five personality theory, metacognition theory, etc. (3) Social Aspects: Unifies demographic and behavioral traits based on social identity theory, behavioral psychology, attachment theory, etc. Due to the lengthy initial personalization tree, we include the full schema in our codebase, where the complete content can be inspected.

\begin{table*}[h]
\centering
\resizebox{\textwidth}{!}{%
\begin{tabular}{p{\textwidth}}
\toprule
\textbf{Prompts for Operational Generation in PersonaTree} \\
\hline
You are a Memory-Tree Operation Generator. You will be given:\\
(1) An initial persona schema represented as a hierarchical JSON tree.\\
(2) A dialogue history.\texttt{\textbackslash n\textbackslash n}\\ 
Your objective is to transform the dialogue history into a sequence of operations for updating the persona schema, **covering as comprehensively as possible all information about this person, especially personalized characteristics**.\texttt{\textbackslash n\textbackslash n}\\

About the schema:\\
- The schema below contains **user attribute information that has already been successfully structured**;\\
- Treat the schema as “recorded information” and **do not re-extract fields that already exist**;\\
- Generate operations for the schema only when the dialogue history introduces additional facts, details, or preferences not yet covered by the schema;\\
- If the dialogue history conflicts with the schema, the **most recent explicit statement** in the dialogue should prevail.\texttt{\textbackslash n\textbackslash n}\\

Principles for using ADD / UPDATE / DELETE / NO\_OP:\\
   * Use: ADD(path, "value") when an attribute at that path has **not been recorded at all**. Prefer creating more branches and avoid overly long content in a single attribute.\\
   * Use: UPDATE(path, "value") when an attribute at that path already has a record and the current passage **supplements, refines, or corrects** it.\\
   * Use: DELETE(path, None) only when the passage explicitly states that an existing piece of information **is no longer valid, is negated, or should be removed**.\\
   * If the passage does not entail any changes, output a single line: NO\_OP().\texttt{\textbackslash n\textbackslash n}\\

Key requirements for "value" in UPDATE (very important):\\
   * "value" must semantically **contain or integrate the previously valid information** while incorporating or reflecting the new information, yielding a more complete, more accurate, and up-to-date description.\\
   * It is **strictly forbidden** to discard useful original content and keep only the new information in an UPDATE.\\
   * When the new information is supplemental or more specific, the value should be an integrated expression of “original information + new supplementation”.\\
   * When the new information conflicts with the old, the value should describe the “current latest and most reasonable state”, while retaining non-conflicting old details whenever possible.\texttt{\textbackslash n\textbackslash n}\\

Notes:\texttt{\textbackslash n}1. Treat each leaf node in the JSON schema as an attribute slot capable of storing a textual value.\texttt{\textbackslash n\textbackslash n}\\

2. For each distinct user personal attribute mentioned in the dialogue history:\\
   * Locate the most closely matching and most specific leaf node in the schema.\\
   * Generate **exactly one and only one** operation for that attribute.\texttt{\textbackslash n\textbackslash n}\\

3. You may use only the following operations:\\
   * ADD(path, "value"), UPDATE(path, "value"), DELETE(path, None), NO\_OP()\texttt{\textbackslash n\textbackslash n}\\

4. Requirements for the "path" format:\\
   * Use a JSON key path separated by English periods. Example:\\
1\_Biological\_Characteristics.Physiological\_Status.Age\_Related\_Characteristics.Chronological\_Age\texttt{\textbackslash n\textbackslash n}\\

5. Requirements for the "value" format:\\
   * Provide a natural-language expression extracted from or normalized based on the dialogue history.\\
   * It must be enclosed in English double quotation marks.\texttt{\textbackslash n\textbackslash n}\\

6. Output format (must be strictly followed):\\
   * Output only operations, one operation per line.
   * Do not add any explanations or comments.
   * The only permissible forms are: ADD(<path>, "<value>"), UPDATE(<path>, "<value>"), DELETE(<path>, None), NO\_OP()\texttt{\textbackslash n\textbackslash n}

Persona Schema:\texttt{\textbackslash n\textbackslash n}\{schema\}\texttt{\textbackslash n\textbackslash n}\\

Dialogue History:\texttt{\textbackslash n\textbackslash n}\{dialogue\_text\}\texttt{\textbackslash n\textbackslash n}\\
Now, based on the given dialogue history, output only the operations:\\
\bottomrule
\end{tabular}%
}
\caption{Prompt for operational generation in PersonaTree for training and inference.}
\label{zztab:7}
\end{table*}

\begin{table*}[h]
\centering
\resizebox{\textwidth}{!}{%
\begin{tabular}{p{\textwidth}}
\toprule
\textbf{Reward-Function Prompt} \\
\hline
You are a strict "overall scorer for attribute-tree operations". Your task is to assign an overall quality score in [-1, 1] to the model-predicted operation sequence Pred\_Ops, given the ground-truth annotated operation sequence GT\_Ops.\\\\

[Input]\\
- GT\_Ops (ground truth): a list of operations, where each element is of the form ADD(path, value) / UPDATE(path, value) / DELETE(path, value) / NO\_OP()\\
- Pred\_Ops (prediction): a list of operations in the same format as above\\
\\\\
\text{[}Critical Constraints\text{]}\\
1) Output only a single JSON object: \{"score": <float>\}. Do not output any explanation and do not include any extra fields.\\
2) score must be a continuous floating-point number within [-1, 1] (any value is allowed). It is recommended to keep 2 decimal places.\\
3) The "score-tier reference" below serves only as anchors for aligning overall quality. You should fine-tune between anchors to output a more granular score.\\
4) For example, if the overall quality falls between 0.7 and 1.0, output a value in [0.71, 0.99]; if it falls between 0.5 and 0.7, output a value in [0.51, 0.69]; and so on.\\\\

[Score-Tier Reference (Overall Quality Anchors)]\\
* 1.0 (nearly perfect): Pred and GT are almost entirely consistent on key operations; types/paths are nearly identical; values are semantically equivalent; no redundant operations.\\
* 0.7 (high quality): most key operations are correct; only minor value-level deviations, or very few missing/redundant operations.\\
* 0.5 (moderately usable): the overall approach and core direction are correct; some missing/redundant operations exist; some paths/values are incorrect, but the main semantics are not affected.\\
* 0.3 (partially reliable): about half of the content is reliable; some key operations are correct while others are wrong, requiring some fixes.\\
* 0.0 (slightly correct): only a small number of operations or fragments are correct; missing/redundant operations and errors are evident; key operations are mixed correct/incorrect.\\
* -0.3 (barely relevant): broadly related but with many omissions/errors; it is only apparent that the model is attempting the task, and it is essentially unusable as-is.\\
* -0.5 (clearly off-target): most key operations are missing or incorrect; many wrong paths/types or obviously redundant operations; overall deviates from expectations.\\
* -0.7 (catastrophic): large-scale structural/semantic disorder; almost unusable.\\
* -1.0 (meaningless output): clearly meaningless, garbage text, or unrelated to the task.\\\\

[Output Format]\\
Output only the JSON object containing the score, with no additional notes or explanations.\\
Output only:\\
\{"score": <float>\}
\\\\
\text{[}Task Data\text{]}\\
  - GT\_Ops:\\
  \{gt\_ops\}\\
\\
  - Pred\_Ops:\\
  \{pred\_ops\}\\
\bottomrule
\end{tabular}%
}
\caption{Reward-function prompt for process-reward RL training.}
\label{zztab:8}
\end{table*}

\end{document}